\pdfoutput=1
\documentclass[11pt]{article}

\usepackage[final]{acl}

\usepackage{times}
\usepackage{upgreek}

\usepackage{latexsym}
\usepackage{graphicx}
\usepackage{colortbl}
\usepackage[T1]{fontenc}
\usepackage[utf8]{inputenc}

\usepackage{CJKutf8}
\usepackage{rotating}
\usepackage{multirow}
\usepackage{booktabs}
\usepackage{amsmath} % 导入 amsmath 宏包
\usepackage{bm}
\usepackage{makecell}
\usepackage{graphicx}

\usepackage{microtype}

\usepackage{inconsolata}
\usepackage{graphicx}
\usepackage{makecell}

\usepackage{amsmath}  % 解决正文表格等大规模报错
\newcommand{\etal}{\textit{et al.}}
\newcommand{\etc}{etc.}
\usepackage{amsfonts} 
\usepackage{amssymb}  

\usepackage{authblk}

\title{ContextBLIP: Doubly Contextual Alignment for Contrastive Image \\ Retrieval  from Linguistically Complex Descriptions}

\author[1 *]{Honglin Lin}
\author[1 *]{Siyu Li}
\author[1 $\dagger$]{Guoshun Nan}
\author[1]{\\Chaoyue Tang}
\author[1]{Xueting Wang}
\author[1]{Jingxin Xu}
\author[1]{Yankai Rong}
\author[2]{\\Zhili Zhou}
\author[3]{Yutong Gao}
\author[1]{Qimei Cui}
\author[1]{Xiaofeng Tao}
\affil[1]{Beijing University of Posts and Telecommunications}
\affil[2]{Guangzhou University ~~~~~~$^{3}$Minzu University of China}

\begin{document}

\maketitle

\begin{abstract}
Image retrieval from contextual descriptions (IRCD) aims to identify an image within a set of minimally contrastive candidates based on linguistically complex text. Despite the success of  VLMs, they still significantly lag behind human performance in IRCD. The main challenges lie in aligning key contextual cues in two modalities, where these subtle cues are concealed in tiny areas of multiple contrastive images and within the complex linguistics of textual descriptions. This motivates us to propose ContextBLIP, a simple yet effective method that relies on a doubly contextual alignment scheme for challenging IRCD. Specifically, 1) our model comprises a multi-scale adapter, a matching loss, and a text-guided masking loss. The adapter learns to capture fine-grained visual cues. The two losses enable iterative supervision for the adapter, gradually highlighting the focal patches of a single image to the key textual cues. We term such a way as \textit{intra-contextual alignment}. 2) Then, ContextBLIP further employs an inter-context encoder to learn dependencies among candidates, facilitating alignment between the text to multiple images. We term this step as \textit{inter-contextual alignment}. Consequently, the nuanced cues concealed in each modality can be effectively aligned. Experiments on two benchmarks show the superiority of our method. We observe that ContextBLIP can yield comparable results with GPT-4V, despite involving about 7,500 times fewer parameters. Our code is available at \hyperlink{https://github.com/LHL3341/ContextBLIP}{https://github.com/LHL3341/ContextBLIP}.

\end{abstract} 
\renewcommand{\thefootnote}{}
\footnote{$^*$Equal Contribution.}
\footnote{$^\dagger$Corresponding author.}
\section{Introduction}
\label{sec:intro}

%-------------------------------------------------------------------------
\begin{figure}
\vspace{-4mm}
\includegraphics[width=1.0\columnwidth]{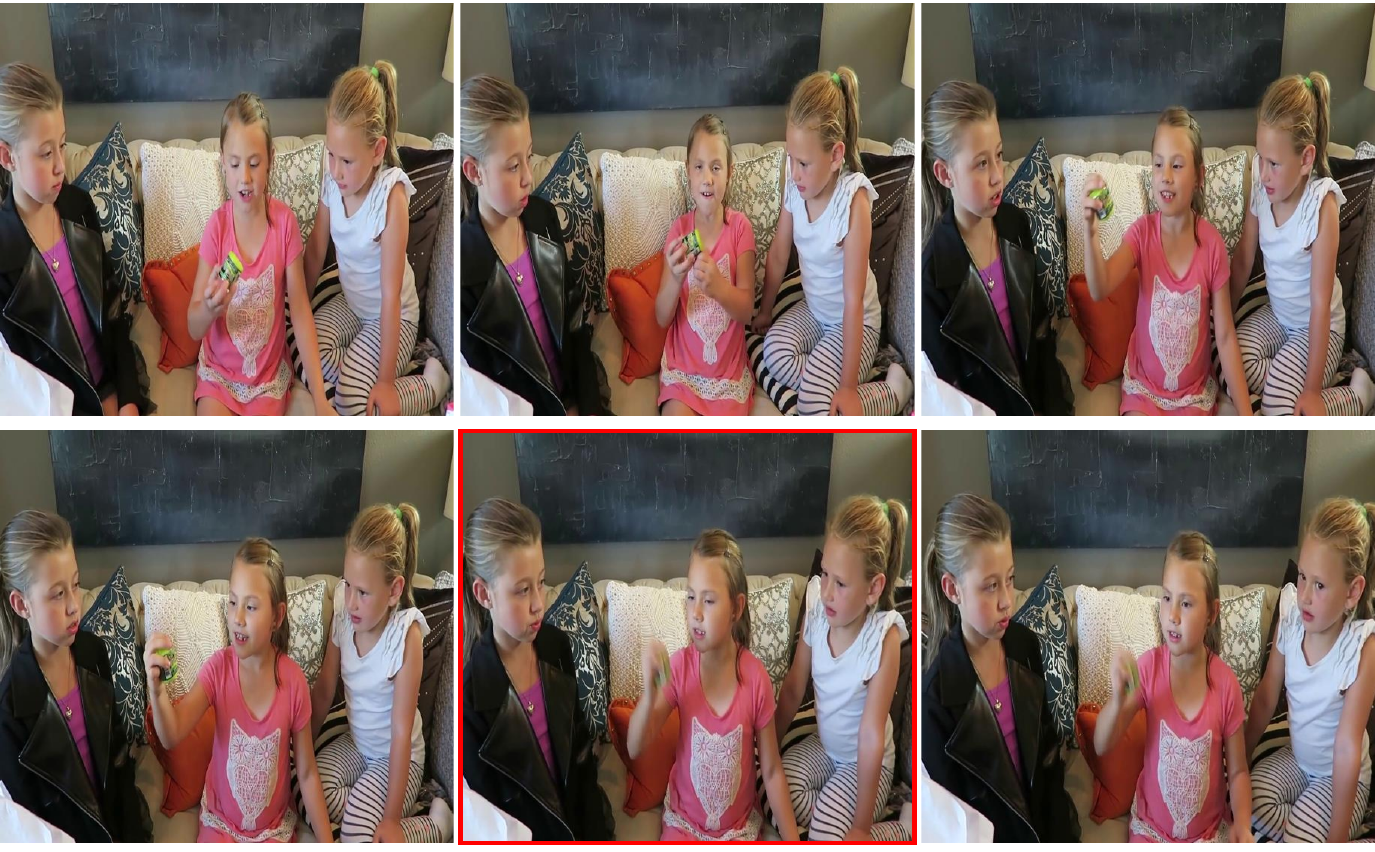} 
\caption{An instance selected from a public benchmark of IRCD, which involves six very similar contrastive image candidates, and the query \textit{``Middle girl's hand is blurry and shoulder level, her eyes are almost shut, the girl on the right is looking at the middle girl's hand''}. The target image is the 4-th one in red rectangular box.}
\label{fig:intro}
\vspace{-6mm}
\end{figure}

%background
Text-to-image retrieval is a fundamental cross-modal task that aims to search images for textual queries. Early studies relied on convolutional neural networks (CNN)~\cite{lecun1998gradient} and Long Short-Term Memory (LSTM)~\cite{10.1162/neco.1997.9.8.1735} to learn to extract image and text features and then proceeded to align the representations of two modalities for retrieval. Recent proliferated large vision-language models (VLMs), such as vision Transformers (ViTs)~\cite{dosovitskiy2021image}, UNITER~\cite{chenUNITERUNiversalImageTExt2020} and CLIP~\cite{radford2021learning}, which are trained on large-scale short text-image corpus, have made remarkable progress for retrieving images from sentences with few objects and simple linguistic. 

However, in the real world, natural languages are highly contextual \cite{formroyal2001,Krojer_2022} with long utterances. Context, including perceptual and temporal cues, plays a pivotal role in grounding the implication of a linguistically complex text~\cite{li2023neural}. Figure \ref{fig:intro} demonstrates such a case that identifies an image from six very similar candidates with a grammatically complicated description. Two major challenges of the retrieval are: 1) A model needs to understand nuanced textual cues, such as \textit{``hand is blurry''}, \textit{``eyes are almost shut''}, and \textit{``looking at...''} across three grammatically complex sentences and align them with various context cues in each image. 2) Long-range dependencies among candidate images need to be captured to perform cross-modal reasoning for further alignment. The above interesting and challenging task is known as image retrieval from contextual descriptions (IRCD)~\cite{Krojer_2022}.

Despite the great success of VLMs, they can hardly tackle the above two challenges, and significantly lag behind human performance. Existing VLMs applied to text-image retrieval mainly include contrastive-based ones such as CLIP~\cite{radford2021learning} and ALIGN ~\cite{jiaScalingVisualVisionLanguage2021}, randomly mask-based ones such as M3AE~\cite{gengMultimodalMaskedAutoencoders2022} and MaskVLM~\cite{kwonMaskedVisionLanguage2023}, and attention-based ones such as ALBEF~\cite{liAlignFuseVision2021} and BLIP~\cite{liBLIPBootstrappingLanguageImage2022}. These models have some limitations. 1) The former two focus more on high-level semantic alignment, while the fine-grained contextual cues may be largely ignored. 2) Existing mask-based ones randomly remove image patches, without specifically learning to concentrate on the key objects associated with text tokens, e.g., ``\textit{hand}'' and ``\textit{eyes}'' in Figure \ref{fig:intro}. 3) Dependencies among images are not specifically considered. 

Previous state-of-the-art NDCR~\cite{li2023neural} proposed for IRCD divided the complex alignment into multiple simple ones and then combined them for final retrieval. However, the performance is highly dependent on the candidate's distributions and is poor for fine-grained alignment on static images with a large variance. We observe that NDCR can hardly capture the key contextual cues in grammatically complex long sentences. Further, it also lacks zero-shot capability. Consequently, NDCR still suffers from the two challenges of IRCD. Details are discussed in Table \ref{tab:main-results-pri} of Experiments.

To this end, we introduce ContextBLIP, a novel doubly contextual alignment scheme for the challenging IRCD task based on BLIP~\cite{liBLIPBootstrappingLanguageImage2022}. %Our ContextBLIP involves two key phases, including intra-contextual and inter-contextual alignments, to learn fine-grained textual and visual interactions and long-range dependencies among contrastive image candidates, respectively. 
1) Specifically, our ContextBLIP comprises a multi-scale adapter, a matching loss, and a text-guided masking loss. The learnable adapter, which is inserted into frozen BLIP, aims to capture higher-level and lower-level visual features of the candidate images. The two losses enable iterative supervision during the training stage, gradually allowing the adapter to highlight the focal patches of a single image to the linguistically complex textual cues. Such a way is termed as \textit{intra-contextual alignment} that aims to tackle the first challenge issue of the IRCD task. 2) Then, we further fine-tune ContextBLIP with a temporal Transformer to learn dependencies among candidate images, facilitating alignment between text to multiple images. This step is termed as \textit{inter-contextual alignment} that aims to address the second challenge issue of IRCD. Experiments on a public benchmark show the effectiveness of the proposed ContextBLIP. The main contributions of this paper are listed as follows. 
\begin{itemize}
    \item We propose ContextBLIP, a simple yet effective method that relies on a doubly contextual alignment scheme for IRCD. It consists of a multi-scale adapter, a matching loss, and a text-guided masking loss, to learn to align the nuanced visual and textual cues, thus effectively tackling the first challenge of IRCD.
    \item We apply ContextBLIP for the zero-shot IRCD, and further fine-tune ContextBLIP with a temporal Transformer to learn the dependencies among different candidates, thus properly addressing the second challenge of IRCD.  
    \item We conduct extensive experiments under various settings to show the superiority of our method. Our ContextBLIP can achieve comparable performance with proliferated GPT-4V~\cite{yang2023dawn} under various prompts. We also evaluate our ContextBLIP on a very recent benchmark MMVP-VLM and the results further confirm the effectiveness of the proposed method.
\end{itemize}

\section{Related Work}
\begin{figure*}[!h]
\centering
\includegraphics[width=2.0\columnwidth]{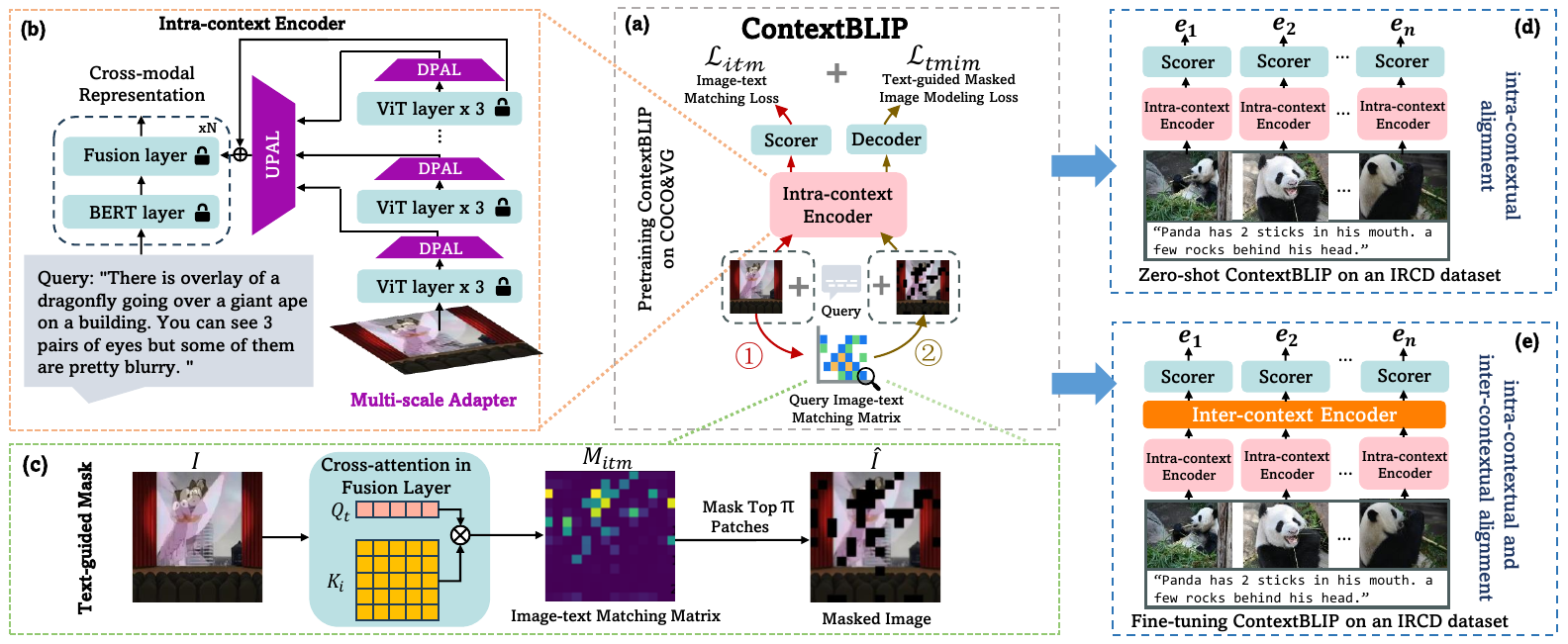} 
    \caption{(a) Architecture of our ContextBLIP, including a BLIP-based {\textbf{\color{pink}intra-context encoder}}, a scorer for image-text matching (ITM, $\mathcal{L}_{itm}$), and a Transformer-based decoder for text-guided masked image modeling (TMIM, $\mathcal{L}_{tmim}$). (b) The {\textbf{\color{violet}{multi-scale adapter}}} in the encoder is co-supervised by $\mathcal{L}_{itm}$ and $\mathcal{L}_{tmim}$ on COCO\&VG datasets, while BLIP is frozen. (c) The learnable {\color{green}{text-guided mask}} is iteratively updated under the co-supervision. (d) Zero-shot ContextBLIP on the IRCD task. (e) Fine-tuning ContextBLIP for IRCD with the {\textbf{\color{orange}inter-context encoder}}.} 
% and (c) (d) the diagram of proposed methods, featuring multi-scale adapter and cross-modal masking. 
%It consists of multiple compressors and an integrator. We insert compressors into different layers of the visual encoder to obtain multi-scale visual features. The compressors project the multi-scale visual features into a shared space. The concatenated multi-scale visual features are then fed into the Integrator, which produces the final representation. 
\label{fig:arc}
\end{figure*}
\textbf{Vision-language models (VLMs):} VLMs~\cite{radford2021learning,liAlignFuseVision2021,wangOFAUnifyingArchitectures2022,liBLIPBootstrappingLanguageImage2022,li2023blip2,zhai2023sigmoid,fang2023data,xu2023demystifying,sun2023evaclip} have shown great potential on image-text retrieval~\cite{lin2015microsoft,krishna2016visual,plummer2016flickr30k}. However, tuning these models for new tasks is expensive. Adapters~\cite{houlsbyParameterEfficientTransferLearning2019,gaoCLIPAdapterBetterVisionLanguage2021,zhangTipAdapterTrainingfreeCLIPAdapter2021,chen2023vision,lu2023uniadapter}, which can be inserted into a pre-trained VLM model, facilitate efficient adaptation to new tasks. The key difference between our adapter and previous ones: ours comprises multiple down-projection adapter layer (DPAL) that connect to the same up-projection adapter layer (UPAL). Each DPAL is inserted into distinct VLM layers, while the UPAL resides outside of the VLM. Such a design can effectively capture multi-level subtle cues. While previous ones include multiple pairs of DPAL and UPAL inserted in VLM.  
%considerably
%~\cite{parkHowVisionTransformers2022}

%Our approach is different from previous works on: 1) Instead of fine-tuning on downstream datasets, we use a small part of the pre-training dataset with lightweight adapters for stronger cross-modal alignment through self-supervised training. 2) We employ a laconic "compress-concatenate-aggregate" way to extract multi-scale visual features~\cite{parkHowVisionTransformers2022}, fused across modalities by cross-attention.

\noindent
\textbf{Masked Image Modeling (MIM):} MIM, such as ViBERT~\cite{luViLBERTPretrainingTaskAgnostic2019} and MAE~\cite{he2021masked}, refers to predicting the missing pixels in an image by using the surrounding pixels as context.
MIM has been applied for various tasks, such as robust learning~\cite{wangMaskedImageModeling2023} and generation target~\cite{bao2022beit,baoVLBEiTGenerativeVisionLanguage2022}. Recent studies, such as MaskVLM~\cite{kwonMaskedVisionLanguage2023},  M3AE~\cite{gengMultimodalMaskedAutoencoders2022}, VL-BEiT~\cite{baoVLBEiTGenerativeVisionLanguage2022} extend MIM for both image and language masking. The key difference between ours and existing ones is: we employ a text-guided masking scheme that learns to generate masks under iterative supervision with two losses, thus gradually highlighting the focal patches of a single image to the key textual cues. While the previous ones randomly remove image patches, without specifically learning to focus the key visual objects associated with text tokens.

\section{Our ContextBLIP}

\subsection{Overall Architecture}

%An overview of the overall model architecture is shown in Figure \ref{lab:figure2}. The visual encoder, using the frozen BLIP image encoder, extracts features from input images. The text encoder processes raw text and visual representations, computing multi-modal representations by employing a text-to-image cross-attention mechanism. These multi-modal representations are then passed through a linear classifier, which determines the matching scores.
Figure \ref{fig:arc}(a) 
presents the overall architecture of the proposed ContextBLIP for the challenging IRCD task, which consists of an intra-context encoder based on frozen BLIP, a multilayer perceptron (MLP) module, and an image decoder. 
The overall procedure can be described in three steps.  \textbf{(1) Pretraining:} We first pre-train the proposed ContextBLIP on the large-scale COCO ~\cite{lin2015microsoft} and VG~\cite{krishna2016visual} datasets. Aiming at tackling the first challenge of IRCD, ContextBLIP includes three key ingredients, i.e., a multi-scale adapter (Figure \ref{fig:arc}(b)), and two co-supervision losses including image-text matching (ITM, $\mathcal{L}_{itm}$) loss and text-guided masked image modeling (TMIM, $\mathcal{L}_{tmim}$) loss. \textbf{(2) Intra-context Alignment:} We directly apply pre-trained ContextBLIP for zero-shot IRCD (Figure \ref{fig:arc}(d)) on the public benchmark (e.g., IMAGECODE), effectively aligning the nuanced visual and textual cues. \textbf{(3) Intra- and Inter-context Alignment:} Then, we further fine-tune ContextBLIP (Figure \ref{fig:arc}(e)) on an IRCD benchmark with an inter-context encoder to learn long-range dependencies among image candidates, thus effectively addressing the second challenge of IRCD. Next, we detail four key components of our method, including the multi-scale adapter, $\mathcal{L}_{itm}$, $\mathcal{L}_{tmim}$ and the inter-context encoder.
\subsection{Multi-Scale Adapter}

%While low-level vision features are important for image modeling, BLIP only leverages representations from the top layer of the vision encoder. To enable more fine-grained interaction between vision and language, we design a multi-scale adapter that aligns key context between image and text.

%{\color{red}This is inspired by previous findings~\cite{amir2022deep} that higher layers of ViT tend to capture global and semantic information, and the lower layers excel at encoding fine-grained visual cues. Then we introduce a text-guided masking mechanism--to be done}  

% BLIP utilizes top-layer vision encoder representations, but to enhance vision-language interaction, we introduce a multi-scale adapter for finer context alignment between image and text.
Our multi-scale adapter, which resides in the intra-context encoder of ContextBLIP, aims to learn to align nuanced visual and textual cues from tiny areas of an image and within linguistically complex descriptions, respectively. Figure \ref{fig:arc} (b) illustrates the architecture of the proposed adapter. It comprises multiple down-projection adapter layer (DPAL) and a up-projection adapter layer (UPAL). DPALs are inserted into different ViT layers of frozen vanilla BLIP, while the UPAL resides outside of ViT layers. All DPALs connect to the same UPAL, such that both higher-level and lower-level features of candidate images can be effectively captured. We give the detailed formulation as follows.

For a pair of image-text $(I, T)$, $I \in \mathbb{R}^{h\times w\times c}$, $T \in \mathbb{R}^{1\times t}$, where $h, w, c$ are height, width and the number of channels of the image, $t$ is the number of tokens in sentence $T$. We split an image into $p \times p$ patches and then augment them with positional encoding. We feed these image patches into the $m$-layer ViT. The intermediate visual representations generated at the $l$-th layer of the ViT can be denoted as $X^l$ = $[x^l_1, \cdots, x^l_i, \cdots, x^l_{p^2}]$, where $l \in [1,m], i \in [1, p^2]$ and $x^l_i \in \mathbb{R}^{d}$. Here $d$ indicates the representation dimension of the ViT at the $l$-th layer. We feed $X^l$ into a down-project adapter layer (DPAL) for mapping to lower-dimensional space. The output of DPAL can be expressed as follows,   
\begin{align} 
  & \Tilde{X}^{l} = ~ \mathrm{DPAL}(X^{l}), 
\end{align}
where DPAL is an MLP network and $\Tilde{X}^{l} \in \mathbb{R}^{p^2 \times \Tilde{d}}$. Here $\Tilde{d} = d / \delta$, where $\delta$ is the downsampling rate of DPAL in our adapter. We use the same rate for all DPALs and obtain the output representations from other DPALs. We aggregate these representations by simply concatenating or adding. Then, the aggregated representations $\Tilde{X}$ will be fed into the proposed up-projection adapter layer (UPAL) for up-projection mapping. The output of the UPAL can be expressed as follows,    
\begin{align} 
  & Y = ~ \mathrm{UPAL}(\Tilde{X}), 
\end{align}
where $Y \in \mathbb{R}^{p^2 \times d}$ and UPAL is an MLP network. 

Finally, we add the output of ViT to $Y$ to obtain the final representations of the image for cross-modal matching. The textual query is encoded by frozen BERT~\cite{devlin2019bert} of vanilla BLIP and then is fed into the fusion layer. We feed the representations of the fusion layer's output to the scorer of the intra-context encoder to get a matching score $e$. By doing so, our multi-scale adapter can facilitate fine-grained interactions between subtle visual regions and linguistic concepts. %thereby enhancing the model's ability to synchronize multi-scale subtle visual features with intricate textual descriptions. %This optimization significantly improves the efficiency and accuracy of the image retrieval process. 

\subsection{Co-supervision under $\mathcal{L}_{itm}$ and $\mathcal{L}_{tmim}$}
We train the proposed adapter with two losses, i,e., $\mathcal{L}_{itm}$ and $\mathcal{L}_{tmim}$, that offer collaborative supervision to highlight key contextual cues in two modalities. Our ContextBLIP performs two separate forward computations to calculate $\mathcal{L}_{itm}$ and $\mathcal{L}_{tmim}$. We detail them as follows. 

\noindent
\textbf{Step 1: Computing $\mathcal{L}_{itm}$:} We sequentially feed a pair of image-text into the intra-context encoder and the MLP-based scorer, and obtain the matching score for the pair. The matching loss  $\mathcal{L}_{itm}$ can be expressed as follows.
\begin{equation}
    \mathcal{L}_{itm}=\frac{1}{3N}\sum_{i=1}^{3N}\mathrm{CrossEntropy}(e_i,q_i), 
\end{equation}
where $e_i \in \mathbb{R}^2, i \in [1,3N],$ indicates the matching score of the $i$-th image-text pair, and $q_i \in \mathbb{R}^2, i \in [1,3N],$ refers to the groundtruth label that consists of $0$ and $1$. Here $N$ is the training batch size. Inspired by vanilla BLIP, we additionally generate $2N$ pairs of negative samples based on cosine distances to $N$ pairs. By distinguishing much more similar image candidates, our ContextBLIP can learn to align the nuanced textual and visual context concealed in tiny areas and within complex descriptions. 

\noindent
\textbf{Step 2: Generating masks and computing $\mathcal{L}_{tmim}$:} We rely on cross-attentions outputted by the fusion layer to generate the text-guided image masking matrix. Figure \ref{fig:arc} (c) demonstrates the generation procedure. Specifically, we manually define a mask ratio $\uppi, \uppi \in [0, 1]$, to determine the number of patches to be masked. Top $\uppi$ patches with the highest attention scores will be masked and then we can get the masking matrix. We remove the patches according to the mask matrix, feed the masked image to the intra-context encoder, and then use a Transformer-based decoder \footnote{Details are available in Appendix \ref{apx:model}.} to reconstruct the image. The pixel-level reconstruction loss $\mathcal{L}_{tmim}$ based on mean squared error (MSE) for a $N$-size training batch can be expressed as follows.

\begin{equation}
    \mathcal{L}_{tmim}=\frac{1}{N}\frac{1}{\mu}\sum_{i=1}^{N}\sum_{j=1}^{\mu} \sum_{s=1}^{S}{\mathrm{MSE}(y_{ijs},\hat{y}_{ijs})}, 
\end{equation}
where $\mu$ is the number of masked patches and $S$ is the number of pixels in each patch. Here $\hat{y}_{ijs}$ and $y_{ijs}$ refer to the original and corresponding reconstructed pixel respectively for the $s$-th pixel of the $j$-th masked patch in $i$-th instance of a training batch. 

\noindent
\textbf{Step 3: Iterative refinement:} The total loss of our ContextBLIP can be formulated as:
\begin{equation}
\mathcal{L} = \mathcal{L}_{itm}+\mathcal{L}_{tmim}.
\label{eq:important}
\end{equation}
We iteratively perform the above steps supervised by the two losses to learn to update the parameters of the multi-scale adapter, and refine the text-guided matrix, allowing the proposed ContextBLIP to gradually concentrate on the focal visual contextual keys associated with the textual cues.

\subsection{Inter-context Encoder}
The task of image retrieval from contextual descriptions (IRCD) requires understanding long-range contextual dependencies among candidate images. Keeping this in mind, we introduce a simple yet effective inter-context encoder, which aims to capture rich interactions between candidate images, as well as contextual alignment between a textual query to multiple images. We employ a two-layer Transformer that stacks on top of the intra-context encoder. The underlying design principle is general, and more advanced encoders can be used here for inter-context alignment. Thus, we can effectively tackle the second challenging issue of IRCD. 
\section{Experiments}

\subsection{Experimental Settings}

%The IMAGECODE dataset~\cite{Krojer_2022}, introduced by Krojer \etal. consists of 94,020 images and 21,202 captions. The task requires models to retrieve the matching image for a caption from a set of ten similar images. The dataset contains two subsets: static images and highly challenging video images, where frames in the latter subset are extracted from the same video clip.
We conduct experiments on four datasets, including large-scale COCO ~\cite{lin2015microsoft} and Visual Genome (VG)~\cite{krishna2016visual} for pre-training, IMAGECODE~\cite{Krojer_2022} for zero-shot and fine-turning, and MMVP-VLM\cite{tong2024eyes} for evaluating our fine-tuned ContextBLIP. We pre-train ContextBLIP on $4$ $\times$ A$100$ GPU cards, and other experiments on a RTX3090 GPU card. During the pre-training stage, we configure adapter downsampling rate  $\delta$ as $2$, the mask ratio $\pi$ as $0.25$. We use vanilla BLIP-129M checkpoint as our backbone, which involves 223M parameters. We implement our model on the PyTorch platform\footnote{More hyperparameters are available in Appendix \ref{apx:1}}.

We select eight strong baselines including CLIP~\cite{radford2021learning}, UNITER~\cite{chenUNITERUNiversalImageTExt2020}, ViBERT~\cite{luViLBERTPretrainingTaskAgnostic2019}, OFA~\cite{wangOFAUnifyingArchitectures2022}, ALBEF~\cite{liAlignFuseVision2021}, BLIP~\cite{liBLIPBootstrappingLanguageImage2022}, BLIP-2~\cite{li2023blip2}, and NDCR~\cite{li2023neural}. Note that NDCR is the previous state-of-the-art method on IMAGECODE. We follow the previous work~\cite{Krojer_2022} to use accuracy as the evaluation metric. The IMAGECODE dataset involves three categorizations in the test set, including ``Video'' which indicates the candidate images are collected from video frames, ``Image'' which represents the ones that are constructed based on static images,  and ``All'' is the hybrid of the above two datasets.

\begin{table}[]
\centering
\small
\centering\setlength{\tabcolsep}{2pt}
\scalebox{1}{
\begin{tabular}{lccccccc}
\toprule
\multicolumn{1}{c}{}       & \textbf{}       & \multicolumn{3}{c}{Zero-shot}          & \multicolumn{3}{c}{Fine-tuned}          \\ \cmidrule(r){3-5}  \cmidrule(r){6-8}
                           Method& Params & All  & Video & Image & All  & Video & Image \\ \midrule
CLIP              & 473M            & 22.4          & 15.6           & 47.8           & 29.9          & 22.0           & 59.8           \\
UNITER            & -               & 19.8          & 13.6           & 42.9           & 25.7          & 19.1           & 50.5           \\
ViLBERT           & -               & 19.3          & 13.5           & 40.8           & 24.5          & 18.0           & 49.3           \\
OFA$^\dagger$               & -               & -             & -              & -              & 27.2          & 21.0           & 52.1           \\
ALBEF$^\dagger$             & -               & 27.7          & 15.7           & 73.3           & -             & -              & -              \\
BLIP$^\dagger$              & 223M            & 28.1          & 15.9           & 74.4           & \underline{34.1}          & 22.7           & \underline{77.4}           \\
BLIP-2$^\dagger$           & 1.2B            & \underline{29.4}          & \underline{16.3}           & \textbf{79.2}  & -             & \textbf{-}     & \textbf{-}     \\
NDCR              & 440M            & \textbf{-}    & \textbf{-}     & -              & \underline{34.1}          & \textbf{26.1}  & 64.3           \\ \midrule
Ours              & 240M            & \textbf{31.0} & \textbf{18.8}  & \underline{77.1}           & \textbf{35.7}          & \underline{24.4}           & \textbf{78.5}  \\
\midrule
Human & \multicolumn{7}{c}{\textbf{90.8}}                                                                                   \\ \bottomrule
\end{tabular}}
\caption{Comparisons on IMAGECODE. Our model achieves state-of-the-art accuracy, with only 2.4M more parameters on vanilla BLIP. Baselines marked with $^\dagger$ indicate that we reproduced the scores as no results are publicly available. The best and second-best results are highlighted with \textbf{bold} and \underline{underline}, respectively.}
\label{tab:main-results-pri}
\end{table}

\subsection{Main Results}

\noindent
\textbf{Zero-shot on IMAGECODE:}
We per-train the proposed ContextBLIP on COCO and VG. Table \ref{tab:main-results-pri} reports the comparisons between ours and the baselines for zero-shot ContextBLIP on the IMAGECODE dataset. Equipped with our multi-scale adapter that only involves 2.4M parameters, our ContextBLIP achieves state-of-the-art performance on all test instances and video frames. Compared to the existing CLIP that uses global information in text and images for alignment, our ContextBLIP obtains $8.6\%$ higher accuracy. Our method also outperforms existing UNITER and VILBERT by $10.2\%$ and $10.7\%$, respectively. The two methods employ random masks for cross-modal alignment. We attribute the improvement to \textit{intra-context alignment} based on multi-scale adapter and text-guided masking. The former learns both higher- and lower-level visual features, and their rich interactions at each level. The latter enables our ContextBLIP to concentrate on focal contextual cues. We also observe that existing BLIP-2 performs better than ours by $2.1$ points on static images, and this is not a surprise as BLIP-2 is $50$ larger than ContextBLIP.

\begin{table*}[!h]
\centering
% \scriptsize
\small
\centering\setlength{\tabcolsep}{2pt}
\begin{tabular}{lcccccccc}
\toprule
\textbf{}               & Context & Quantities & Spatial & Negation & Occlusion & Nuances & Co-reference & Meta Properties \\ \midrule
CLIP (Zero-shot) & 13.3                           & 23.7                & 20.6                       & 11.7              & 14.1                            & 9.9              & 25.3                  & 29.2                     \\
BLIP (Zero-shot) & 15.4                           & \underline{30.9}                & 27.7                       & 11.4              & 16.7                            & 11.4             & 28.9                  & 12.5                     \\
Ours (Zero-shot) & 17.3                           & \textbf{39.2}       & \textbf{36.9}              & \underline{19.0}              & \underline{19.1}                            & 11.0             & \textbf{37.4}         & 25.0                     \\
 \midrule
CLIP (Fine-tuned)  & \underline{19.2}                  & \underline{30.9}                & 30.5                       & 17.3              & 18.6                            & \underline{14.8}             & 32.5                  & \underline{33.3}                     \\
Ours (Fine-tuned)   & \textbf{25.1}                  & \textbf{39.2}                & \underline{31.9}                       & \textbf{25.2}     & \textbf{23.7}                   & \textbf{19.7}    & \underline{36.1}                  & \textbf{37.5}            \\ \bottomrule
\end{tabular}

\caption{Comparison of challenging samples in the IMAGECODE benchmark under zero-shot and fine-tuned settings. The samples involve challenging contextual alignment such as ``Context'', ``Nuances'' and ``Co-reference''. }
\label{tab:main-challenge-samples}
\end{table*}

\begin{table*}[]
\centering
\small
\centering\setlength{\tabcolsep}{2pt}
\footnotesize
\begin{tabular}{l|cc|ccccccccc|c}
\toprule
                   & \makecell[c]{Image\\Size} & Params  & \includegraphics[width=0.03\textwidth]{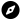}    & \includegraphics[width=0.03\textwidth]{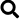}    & \includegraphics[width=0.03\textwidth]{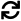}    & \includegraphics[width=0.03\textwidth]{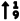}   & \includegraphics[width=0.018\textwidth]{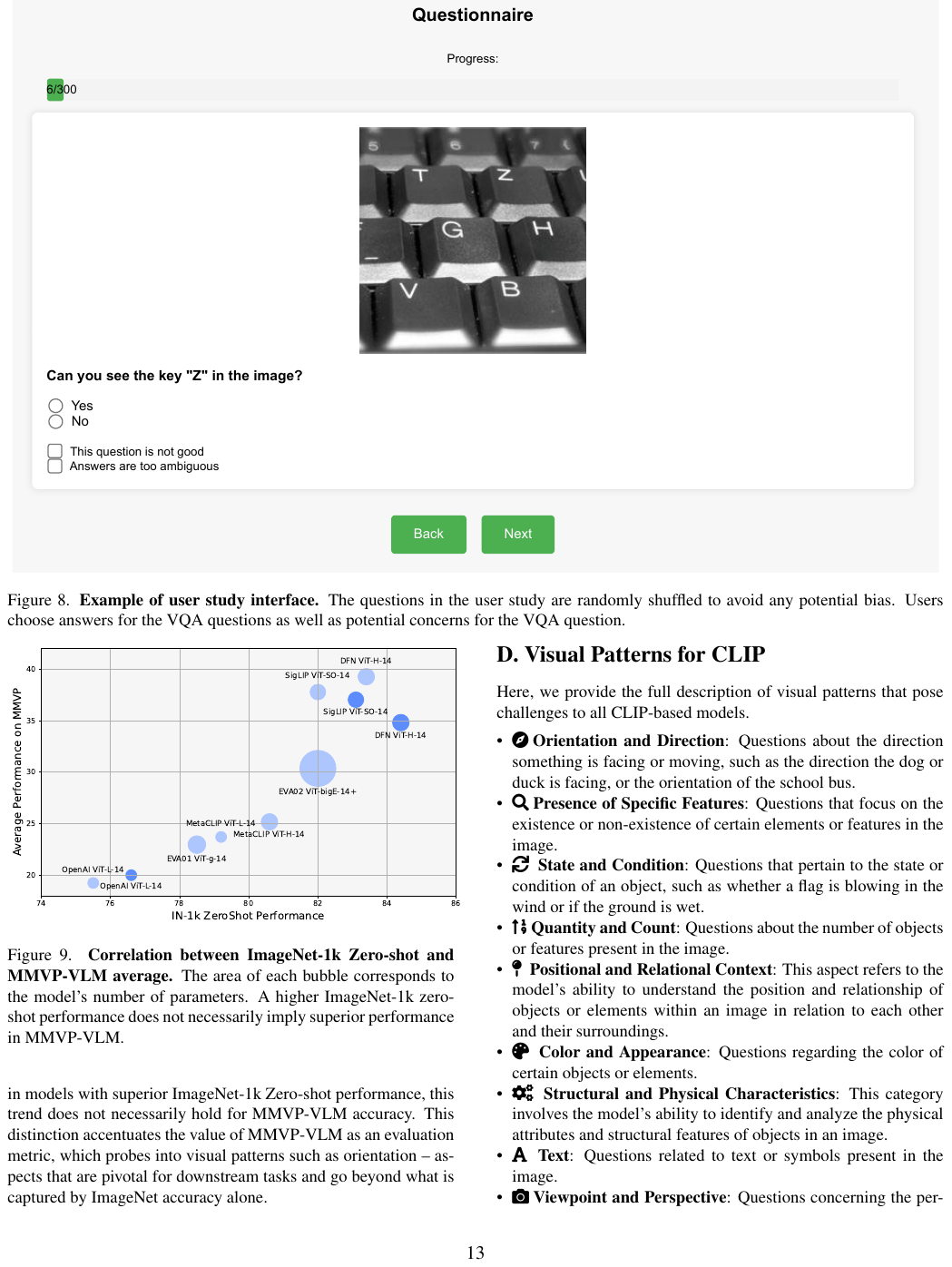}    & \includegraphics[width=0.03\textwidth]{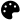}    & \includegraphics[width=0.04\textwidth]{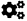}    & \includegraphics[width=0.03\textwidth]{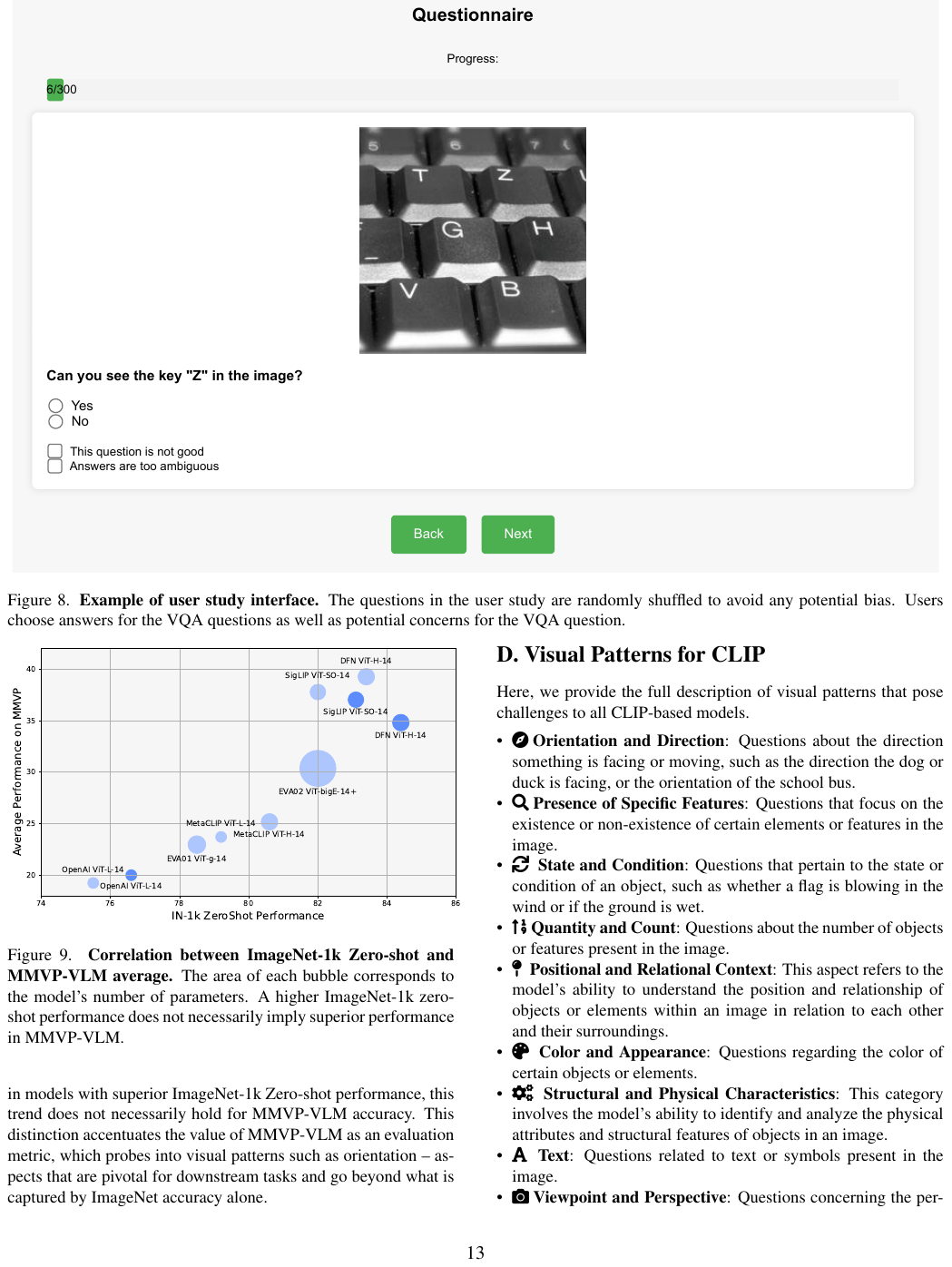}    & \includegraphics[width=0.03\textwidth]{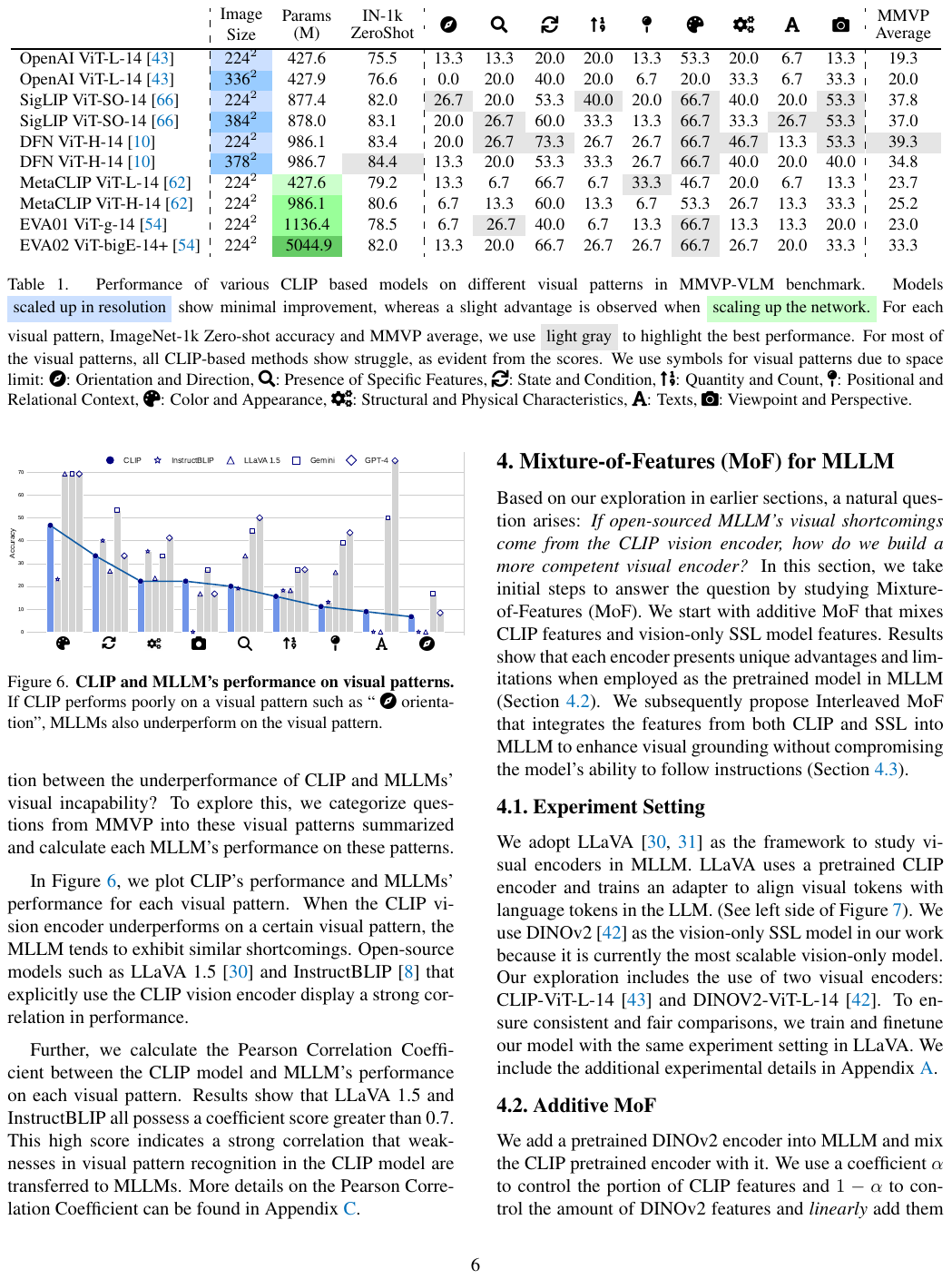}    & \makecell[c]{MMVP\\Average} \\ \midrule

\makecell[c]{DFN ViT-H-14\\\cite{fang2023data}}       & $224^2$                                       & 986.1M  & \underline{20.0} & \underline{26.7} & \textbf{73.3} & \underline{26.7} & \underline{26.7} & \underline{66.7} & \textbf{46.7} & \underline{13.3} & \textbf{53.3} & \underline{39.3}                                        \\
\makecell[c]{MetaCLIP ViT-H-14\\\cite{xu2023demystifying}}  & $224^2$                                       & 986.1M  & 6.7  & 13.3 & 60.0 & 13.3 & 6.7  & 53.3 & 26.7 & \underline{13.3} & 33.3 & 25.2                                        \\
\makecell[c]{EVA02 ViT-bigE-14+\\\cite{sun2023evaclip}} & $224^2$                                       & 5044.9M & 13.3 & 20.0 & \underline{66.7} & 13.3 & \underline{26.7} & \underline{66.7} & 26.7 & \textbf{20.0} & 33.3 & 33.3                                        \\
BLIP~\cite{liBLIPBootstrappingLanguageImage2022} &$224^2$  & 223M & 13.3 & 6.7 & 40.0 & 20.0 &\underline{26.7} & \underline{66.7} & \textbf{46.7} & \textbf{20.0} & \underline{46.7} & 31.9 \\ \midrule
ContextBLIP (Ours)               & $224^2$                                       & \textbf{240M}    & \textbf{26.7} & \textbf{46.7}  & 60.0 & \textbf{40.0} & \textbf{46.7} & \textbf{93.3} & \underline{40.0} & \textbf{20.0} & \textbf{53.3} & \textbf{47.4}    \\ \bottomrule

\end{tabular}
\caption{Comparisons on challenging visual patterns on the MMVP-VLM benchmark. We follow the previous work~\cite{tong2024eyes} to use the symbols, such as \includegraphics[width=0.05\columnwidth]{imgs//table/1.pdf} and \includegraphics[width=0.05\columnwidth]{imgs//table/2.pdf}, to indicate $9$ challenging visual patterns. More details can be found in Appendix \ref{apx:B}. The best and second-best results are marked in \textbf{bold} and \underline{underline}.}
\label{tab:main-MMVP}
\vspace{-5mm}
\end{table*}

\noindent
\textbf{Fine-tuned on IMAGECODE:}
The right side of Table \ref{tab:main-results-pri} reports the comparisons of our fine-tuned ContextBLIP to baselines.
Our ContextBLIP can achieve state-of-the-art accuracy on test sets of ``All'' and ``Image'' and this further confirms the superiority of our method. ContextBLIP outperforms exiting CLIP and OFA by $18.7$ and $26.4$ points on static images, showing the effectiveness of both intra-context and inter-context alignment. We also observe that the proposed ContextBLIP performs slightly worse than the previous state-of-the-art NDCR by $1.7$ points on video frames. However, NDCR pays more attention to the cross-modal alignment of video frames and complex text, while the ability of static images is largely underexplored. We observe that ours is better than NDCR by $14.2$ points on ``Image'', with nearly half fewer parameters. We also attribute this to the model's capability to handle distributions of the candidate image, where static ones present a large variance.

\noindent
\textbf{Comaprison of challenging samples of IMAGECODE:}
\label{sec:task-specific}
We compare our method with the existing CLIP and BLIP on the $200$ challenging samples highlighted in IMAGECODE. These samples are manually labeled with high-quality annotations. Table \ref{tab:main-challenge-samples} shows that our ContextBLIP consistently performs best under the fine-tuned setting on various scenarios, and performs best on most scenarios under the zero-shot setting. These results further confirm the superiority of our ContextBLIP in tackling the two challenges of the IRCD task.%  discussed at the beginning. %More details of these samples can be found in IMAGECODE. 

\noindent
\textbf{Comparisons on MMVP-VLM Benchmark:}
We compare our ContextBLIP with a very recent benchmark MMVP-VLM \cite{tong2024eyes}, which aims to evaluate how well a VLM model handles various challenging visual patterns. These manually defined patterns such as specific features, and positional and relational context, require a model to capture contextual details of visual cues or perform cross-modal reasoning. Table \ref{tab:main-MMVP} shows that our ContextBLIP, which is fine-tuned on the IMAGECODE dataset, achieves the best accuracy on most of the patterns. Further, the proposed ContextBLIP involves much fewer parameters, e.g, the number of parameters in MetaCLIP ViT-H-14 \cite{xu2023demystifying} is nearly $25$ times more than ContextBLIP, while average accuracy is lower than ours by $22.2$ points.  

\begin{table}[!h]
\centering
\small
\centering\setlength{\tabcolsep}{2pt}
\footnotesize
\begin{tabular}{lcc}
\toprule
                                 & Zero-shot & Fine-tuned \\ \midrule
w/o Multi-scale Adapter & 15.9               & 22.7               \\
w/o $\mathcal{L}_{tmim}$ & \underline{18.2}                & 23.0               \\
w/o Inter-context Encoder & - & \underline{23.4}   \\
ContextBLIP (Ours)                    & \textbf{18.8}      & \textbf{24.4}      \\ 
\bottomrule
\end{tabular}
\caption{Ablation study on the IMAGECODE dataset.}
\label{tab:ablation}
\end{table}
\subsection{Ablation Study}
We conduct an ablation study on IMAGECODE to measure the contribution of each component to IRCD. Table \ref{tab:ablation} reports the results. Under both zero-shot and fine-tuned settings, we observe that the removal of the multi-scale adapter leads to a significant performance decrease, i.e., 2.9 points and 1.7 points, indicating the effectiveness of the adapter. Under the fine-tuned setting, it shows that the removal of the inter-context encoder leads to a $1$ point performance drop, suggesting the effectiveness of long-range dependencies for the retrieval.

\iffalse
The results in Table \ref{tab:ablation} show the performance of our method compared to ablated variants in zero-shot and fine-tuned settings on the retrieval task.

In the zero-shot setting, our full model outperforms the model without adapters, demonstrating the benefit of adapter modules for leveraging pre-trained knowledge with minimal parameters. Our method also surpasses the model without Cross-modal Masking, indicating masked image modeling effectively improves cross-modal alignment during self-supervised pre-training.

When fine-tuning is applied, all models improve over the zero-shot results as expected by utilizing downstream data. Still, our approach achieves the best overall performance, suggesting fine-tuning can further optimize the cross-modal associations learned by our self-supervised approach. The gains are more evident in video domains compared to images, showing our method is particularly effective for the more challenging cross-modal matching between language and dynamic visual content in videos.

After fine-tuning with the temporal transformer, the model's performance has been further improved. This demonstrates that our temporal transformer can effectively learn the dependencies among candidate images, facilitating alignment between text and multiple images, thereby addressing challenging tasks like IRCD.

In summary, the ablation study validates the importance of both adapter modules and masked image modeling within our framework for boosting multimodal representation learning from unlabeled data.
\fi
\subsection{Sensitivity Analysis}
We conduct experiments for sensitivity analysis on IMAGECODE under the zero-shot setting.

\noindent
\textbf{Masking ratio:} To evaluate how mask ratio $\pi$ affect the retrieval, we configure $\pi$ as $0.25$, $0.50$ and $0.75$, respectively. Table \ref{tab:maskratio} shows lower masking can increase the accuracy, e.g., from $75.4\%$ to $79.5\%$. This interesting finding on the IRCD task does not align with previous studies that advocated for higher masking ratios~\cite{he2021masked,gengMultimodalMaskedAutoencoders2022,baoVLBEiTGenerativeVisionLanguage2022,kwonMaskedVisionLanguage2023}. One possible underlying reason is the challenging alignment in ICRD require more dense visual cues among similar candidates.  

\begin{table}[!h]
\centering
\small
\centering\setlength{\tabcolsep}{2pt}
\footnotesize
\begin{tabular}{lccc}
\toprule
Mask Ratio & All  & Video & Image \\ \midrule
0.25       & \textbf{31.1} & 19.9           & \textbf{79.5}  \\
0.50        & 31.0          & \textbf{20.6}  & 76.3           \\
0.75       & 29.8          & 19.4           & 75.4           \\ \bottomrule
\end{tabular}
\caption{Sensitivity analysis of the mask ratio $\pi$.}
\label{tab:maskratio}
\end{table}

\begin{table}[]
\centering
\small
\centering\setlength{\tabcolsep}{2pt}
\footnotesize
\begin{tabular}{lcccc}
\toprule
Layer          & All  & Video & Image & Params \\ \midrule
{[}12{]}       & 30.4          & 19.4           & 78.1           & 223.6M                     \\
{[}3,6,9,12{]} & \textbf{31.1} & \textbf{19.9}  & \textbf{79.5}  & 225.4M                     \\
{[}1-12{]}     & 30.5          & 19.8           & 77.0           & 230.1M                     \\ \bottomrule
\end{tabular}
\caption{Sensitivity analysis of the position of adapter.}
\label{tab:adapter}
\end{table}

\noindent
\textbf{Position of adapter inserted:} We analyze the impact of adapters inserted into different layers on the retrieval. We include three cases, i.e., inserting both down- and up-projection layers in the top layer of BLIP, inserting down-projection layers in $3$-th, $6$-th, $9$-th, $12$-th layers, and inserting down-projection layers in each layer. Table \ref{tab:adapter} shows inserting adapters at multiple layers of BCLIP achieved the highest overall accuracy. This suggests there exists a tradeoff for the number of layers to be inserted.%  that the proposed multi-scale adapter can capture fine-grained features, while 
%confirms our hypothesis that multi-level modulating representations at various depths improves cross-modal alignment. 

%Notably, multi-level insertion resulted in substantial gains relative to the top-level-only configuration, with a modest parameter increase. It is demonstrated that modulating representations at multiple depths through multi-level adapter insertion is more effective. Interestingly, beyond a certain number of layers, the model's performance decreased.

\begin{table}[]
\centering
\small
\centering\setlength{\tabcolsep}{2pt}
\footnotesize
\begin{tabular}{lcccc}
\toprule
Type       & 1   & 2           & 4           & 8   \\ \midrule
All    & 30.2 & \textbf{31.1} & 30.8          & 30.1 \\
Video  & 19.6 & 19.9          & \textbf{20.1} & 19.3 \\
Static & 76.5 & \textbf{79.5} & 77.2          & 77.0 \\ \bottomrule
\end{tabular}
\caption{Sensitivity analysis of downsampling rate $\delta$.}
\label{tab:reduction}
\end{table}

\noindent
\textbf{Downsampling rate $\delta$:} We evaluate how $\delta$ affects the retrieval. We configure $\delta$ as $1$, $2$, $4$, and $8$, respectively. Table \ref{tab:reduction} shows that our ContextBLIP achieves the best performance when $\delta$ is set as $2$. The results present a large variance, e.g., $77.0$ for $\delta = 8$ and $79.5$ for $\delta = 2$. This suggests that a small rate $\delta$ leads to better performance. 
%We conducted an ablation experiment varying the dimensionality (D, D/2, D/4, D/8) of the bottleneck layers in our model architecture to evaluate its impact on performance. As shown in Table \ref{tab:reduction}, for the overall IMAGECODE benchmark, a reduction to half the original dimensionality (D/2) yielded the highest accuracy of 31.1\%, outperforming the other settings.

%A similar pattern emerged for the video subset, with D/4 achieving the maximum 20.1\% performance and D/2 indicating a comparable accuracy of 19.9\%. For static images, D/2 again resulted in the best score of 79.5\%. These findings suggest that moderate reductions in the hidden dimension serve to optimize the relationship between modeling power and complexity for our tasks.

%Both excessive expansion and contraction of the representational subspace appear to degrade modeling abilities. Too low a dimension may insufficiently capture the richness of multi-modal associations, while too high a dimension risks overfitting and decreases parameter efficiency. Future work will explore more sophisticated approaches to dynamically determining bottleneck dimensions.

\subsection{Case Study}
\begin{figure}[!ht]
    \centering
    \includegraphics[width=1.0\columnwidth]{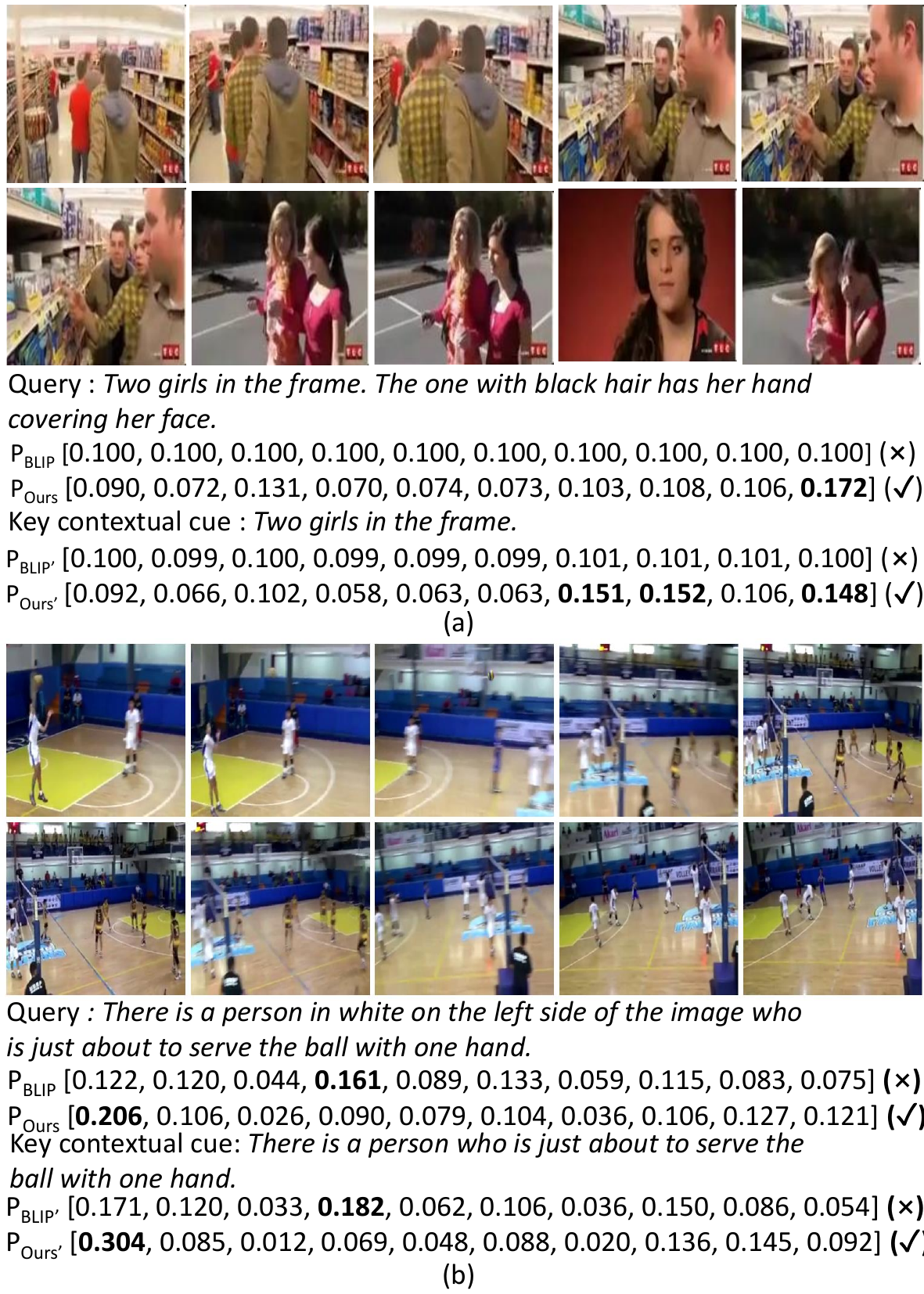}
    \caption{(a) Zero-shot: P\textsubscript{BLIP} and P\textsubscript{Ours} are two matching scores of BLIP and ours, and P\textsubscript{BLIP'}, P\textsubscript{Ours'} are scores for the key contextual cue. (b) Fine-tuned:  P\textsubscript{BLIP} and P\textsubscript{Ours} are two matching scores of BLIP and ours, and P\textsubscript{BLIP'}, P\textsubscript{Ours'} are scores for the key contextual cue.}
    \label{fig:zeroshotcase}
\end{figure}

Figure \ref{fig:zeroshotcase} two cases to visually show why our ContextBLIP performs better for the challenging IRCD task. Both the zero-shot case and fine-tuned case demonstrate that our ContextBLIP not only yields the highest matching score for the golden candidate image. More importantly, it is also capable of aligning the key context cues in two modalities. For example, our method can understand textual cues ``two girls in the frame'' in the long query, and yield more accurate alignment to the $7$-th, $8$-th, and $10$-th candidate images with higher matching scores. Equipped with the proposed inter-context encoder, our ContextBLIP can accurately identify the $1$-st candidate with the highest score.

\subsection{Comparisons with GPT-4V}

% \begin{table}[!ht]
% \centering
% \setlength{\tabcolsep}{2pt}
% \begin{tabular}{lllllll}
% \toprule
%      & \multicolumn{3}{c}{Video}                                          & \multicolumn{3}{c}{Static}                    \\ \hline
%      & 1             & 2             & 3                                  & 1             & 2             & 3             \\ \cline{2-7} 
% GPT4-V & 54\%          & \textbf{50\%} & \multicolumn{1}{l|}{\textbf{46\%}} & 88\%          & \textbf{90\%} & \textbf{94\%} \\
% ContextBLIP (Ours) & \textbf{58\%} & \textbf{50\%} & \multicolumn{1}{l|}{42\%}          & \textbf{90\%} & \textbf{90\%} & 86\%          \\ \bottomrule
% \end{tabular}
% \caption{Comparision with GPT-4V on random samples}
% \label{tab:gpt4}
% \end{table}

\begin{table}[!ht]
\centering
\small 
\centering\setlength{\tabcolsep}{2pt}
\footnotesize
\begin{tabular}{lccc|ccc}
\toprule
& \multicolumn{3}{c}{Video} & \multicolumn{3}{c}{Static} \\
% \cmidrule(lr){2-4} \cmidrule(lr){5-7}
\midrule
& 1 & 2 & 3 & 1 & 2 & 3 \\
\midrule
GPT-4V & \textbf{22\%} & 20\% & 22\% & 46\% & 54\% & 44\% \\
ContextBLIP (Ours) & 20\% & \textbf{24\%} & \textbf{26\%} & \textbf{82\%} & \textbf{80\%} & \textbf{78\%} \\
\bottomrule
\end{tabular}
\caption{Comparison with GPT-4V on random samples.}
\label{tab:comparison_gpt4v}
\end{table}

% \begin{table}[!h]
% \centering
% \small
% \centering\setlength{\tabcolsep}{2pt}
% \footnotesize
% \begin{tabular}{cccc}
% \toprule
% Instance ID & Wrong Matching & Right Matching & All Results \\ \midrule
% 7407       & 0 & 20           & 20  \\
% 7836        & 1          & 19  & 20           \\
% 8129       & 11          & 9           & 20           \\ \bottomrule
% \end{tabular}
% \caption{Statistics of retrieval performance of GPT-4V under $20$ different prompts. ``Instance ID'' indicates the instance number in the IMAGECODE test set.}
% \label{tab:main-various-prompt}
% \end{table}

We compare our ContextBLIP with the OpenAI multi-modal large language model (MLLM) GPT-4V ~\cite{yang2023dawn}. We randomly sample $50$ instances three times from the test set of IMAGECODE, and use a prompt in the form of ``Which image fits the text description best, please output the serial number of the image:$<text\_query>$" to identify the image. Table \ref{tab:comparison_gpt4v} shows that our model can achieve comparable results to GPT-4V, despite involving about 7,500 times fewer parameters. This also aligns with some existing studies ~\cite{tong2024eyes} that GPT-4V may fail to understand subtle cues\footnote{More cases are provided in Appendix \ref{apx:D2} }. Figure \ref{fig:main-gpt-4v} demonstrates how we prompt GPT-4V for the challenging IRCD task. Table \ref{tab:main-various-prompt} reports the matching scores of three instances selected from the test set of IMAGECODE. For the challenging ``ID 8129'', we observe that GTP-4V performs poorly.

\begin{figure}[!ht]
    \centering
    \includegraphics[width=1.0\columnwidth]{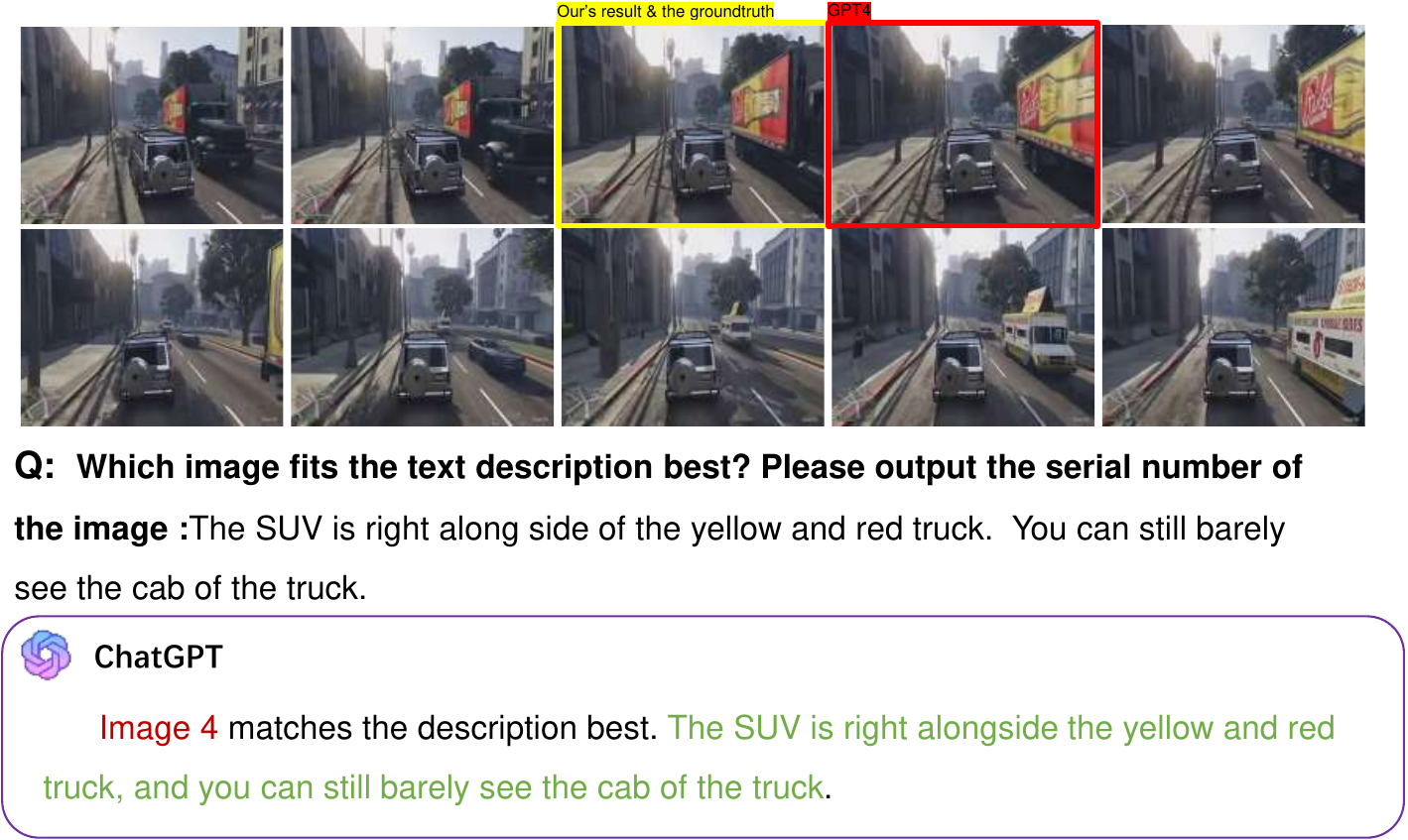}
    \caption{Case illustration of how we prompt GPT-4V for IRCD. The red boxes represent the GPT-4V's response and the yellow one indicates our prediction.}
    \label{fig:main-gpt-4v}
\end{figure}

\iffalse
In addition, we tested the impact of similar meanings prompts on the accuracy of GPT-4V's responses, and the test results are as follows (Table \ref{tab:main-various-prompt}). The results indicate that prompts with similar meanings have little impact on the generated results, indicating that our experiment is reasonable.
\fi

\begin{table}[!h]
\centering
\small
\centering\setlength{\tabcolsep}{2pt}
\footnotesize
\begin{tabular}{cccc}
\toprule
Instance ID & Wrong Matching & Right Matching & All Results \\ \midrule
7407       & 0 & 20           & 20  \\
7836        & 1          & 19  & 20           \\
8129       & 11          & 9           & 20           \\ \bottomrule
\end{tabular}
\caption{Statistics of retrieval performance of GPT-4V under $20$ different prompts. ``Instance ID'' indicates the instance number in the IMAGECODE test set.}
\label{tab:main-various-prompt}
\end{table}

\subsection{More discussion}

We also compare our ContextBLIP with the previous state-of-the-art NDCR \cite{li2023neural} model on the IRCD task. We follow the existing NDCR to divide the linguistic complex descriptions into multiple segments with different lengths. We are interested in such a setting and evaluate how well the proposed ContextBLIP performs over sentences that are split from the same long description. Figure \ref{tab:ndcr}  illustrates that the proposed ContextBLIP consistently outperforms the existing NDCR under various sentence lengths.  
\iffalse
We also compare our ContextBLIP with the previous state-of-the-art NDCR~\cite{li2023neural} model on the IRCD task. We follow the existing NDCR to divide the linguistic complex descriptions into multiple segments with different lengths. We are interested in such a setting and evaluate how well the proposed ContextBLIP performs over sentences that are split from the same long description. Figure \ref{tab:ndcr} illustrates that the proposed ContextBLIP consistently outperforms the existing NDCR. Nonetheless, it struggles to accurately encode all the fine-grained details, particularly with texts containing long and complex propositions, which hinders its ability to fully integrate context and discern key semantic information within the current scene. These challenges highlight areas where ContextBLIP can significantly enhance its capabilities when dealing with the retrieval task of complex text.
\fi

\begin{table}[!h]
\centering
\small
\centering\setlength{\tabcolsep}{2pt}
\begin{tabular}{lcccccc}
\toprule
Nums of props & 1  & 2 & 3 & 4 & 5 & 6\\ \midrule
All & 72 & 899 & 1215 & 99 & 14 & 3 \\
NDCR & 29 & 327 & 384 & 28 & 2 & 0 \\
ContextBLIP & 36 & 391 & 416 & 37 & 3 & 1 \\
\midrule
Improvement (\%) & 9.7 & 7.1 & 2.6 & 9.1 & 7.1 & 33.3 \\
\bottomrule
\end{tabular}
\caption{Comparisons of ContextBLIP with NDCR over various lengths of textual propositions, where $2$ indicates the number of segments split from a long query. }
\label{tab:ndcr}
\end{table}
\vspace{-4mm}
\section{Conclusion}

This paper presents ContextBLIP, a simple yet effective doubly contextual alignment scheme for the challenging IRCD. Our model comprises a multi-scale adapter, a matching loss, and a text-guided masking loss. The adapter learns to capture fine-grained visual cues. The two losses enable iterative supervision for the adapter, gradually highlighting the focal patches of a single image to the key textual cues. Then, ContextBLIP further employs an inter-context encoder to learn dependencies among candidates, facilitating accurate alignment between text to multiple images. Consequently, the nuanced cues concealed in textual and visual modalities can be effectively aligned. Experiments on two benchmarks show the effectiveness of our method. We observe that our ContextBLIP can yield comparable results with GPT-4V, despite involving about 7,500 times fewer parameters. In the future, we plan to extend our method to text-to-video retrieval. 
\section{Limitation}
The adaptive mask ratio is worth considering in the future, as a fixed masking ratio in our paper may not dynamically adapt to different cross-modal interactions. The proposed method may also have limitations for fine-grained retrieval for long videos, as pre-training on long videos is time-expensive and requires very large GPUs.

%In this work, we propose MMBLIP, a new model architecture for efficient vision-language adaption. MMBLIP inserts bottleneck compressors at varied Transformer encoder layers to learn multi-level cross-modal alignments. Additionally, it guides masked regions with cross-attention maps from ITM to provide detailed supervision to help capture fine-grained alignments. The model is able to learn cross-modal alignments at different levels of abstraction through local region reconstruction with explicit supervision, achieving good performance gains while introducing limited parameters. Experiments indicate that our method achieves great progress with few additional parameters, little computational cost, and limited data. Specifically, MMBLIP matched GPT-4V's accuracy on the challenging similar-image-retrieval task.

%Here are some other ways to phrase the potential directions to further enhance MMBLIP's performance:

\bibliography{acl_latex}

\begin{thebibliography}{36}
\expandafter\ifx\csname natexlab\endcsname\relax\def\natexlab#1{#1}\fi

\bibitem[{Bao et~al.(2021)Bao, Dong, Piao, and Wei}]{bao2022beit}
Hangbo Bao, Li~Dong, Songhao Piao, and Furu Wei. 2021.
\newblock Beit: Bert pre-training of image transformers.
\newblock \emph{arXiv preprint arXiv:2106.08254}.

\bibitem[{Bao et~al.(2022)Bao, Wang, Dong, and Wei}]{baoVLBEiTGenerativeVisionLanguage2022}
Hangbo Bao, Wenhui Wang, Li~Dong, and Furu Wei. 2022.
\newblock Vl-beit: Generative vision-language pretraining.
\newblock \emph{arXiv preprint arXiv:2206.01127}.

\bibitem[{Chen et~al.(2020)Chen, Li, Yu, El~Kholy, Ahmed, Gan, Cheng, and Liu}]{chenUNITERUNiversalImageTExt2020}
Yen-Chun Chen, Linjie Li, Licheng Yu, Ahmed El~Kholy, Faisal Ahmed, Zhe Gan, Yu~Cheng, and Jingjing Liu. 2020.
\newblock Uniter: Universal image-text representation learning.
\newblock In \emph{ECCV}, pages 104--120.

\bibitem[{Chen et~al.(2022)Chen, Duan, Wang, He, Lu, Dai, and Qiao}]{chen2023vision}
Zhe Chen, Yuchen Duan, Wenhai Wang, Junjun He, Tong Lu, Jifeng Dai, and Yu~Qiao. 2022.
\newblock Vision transformer adapter for dense predictions.
\newblock \emph{arXiv preprint arXiv:2205.08534}.

\bibitem[{Devlin et~al.(2019)Devlin, Chang, Lee, and Toutanova}]{devlin2019bert}
Jacob Devlin, Ming-Wei Chang, Kenton Lee, and Kristina Toutanova. 2019.
\newblock \href {http://arxiv.org/abs/1810.04805} {Bert: Pre-training of deep bidirectional transformers for language understanding}.

\bibitem[{Dosovitskiy et~al.(2020)Dosovitskiy, Beyer, Kolesnikov, Weissenborn, Zhai, Unterthiner, Dehghani, Minderer, Heigold, Gelly et~al.}]{dosovitskiy2021image}
Alexey Dosovitskiy, Lucas Beyer, Alexander Kolesnikov, Dirk Weissenborn, Xiaohua Zhai, Thomas Unterthiner, Mostafa Dehghani, Matthias Minderer, Georg Heigold, Sylvain Gelly, et~al. 2020.
\newblock An image is worth 16x16 words: Transformers for image recognition at scale.
\newblock \emph{arXiv preprint arXiv:2010.11929}.

\bibitem[{Fang et~al.(2023)Fang, Jose, Jain, Schmidt, Toshev, and Shankar}]{fang2023data}
Alex Fang, Albin~Madappally Jose, Amit Jain, Ludwig Schmidt, Alexander Toshev, and Vaishaal Shankar. 2023.
\newblock \href {http://arxiv.org/abs/2309.17425} {Data filtering networks}.

\bibitem[{Fodor(2001)}]{formroyal2001}
Jerry Fodor. 2001.
\newblock \href {https://doi.org/10.1017/S1358246100010808} {Language, thought and compositionality}.
\newblock \emph{Royal Institute of Philosophy Supplement}, 48:227–242.

\bibitem[{Gao et~al.(2021)Gao, Geng, Zhang, Ma, Fang, Zhang, Li, and Qiao}]{gaoCLIPAdapterBetterVisionLanguage2021}
Peng Gao, Shijie Geng, Renrui Zhang, Teli Ma, Rongyao Fang, Yongfeng Zhang, Hongsheng Li, and Yu~Jiao Qiao. 2021.
\newblock Clip-adapter: Better vision-language models with feature adapters.
\newblock \emph{ArXiv}, abs/2110.04544.

\bibitem[{Geng et~al.(2022)Geng, Liu, Lee, Schuurmans, Levine, and Abbeel}]{gengMultimodalMaskedAutoencoders2022}
Xinyang Geng, Hao Liu, Lisa Lee, Dale Schuurmans, Sergey Levine, and Pieter Abbeel. 2022.
\newblock Multimodal masked autoencoders learn transferable representations.
\newblock \emph{arXiv preprint arXiv:2205.14204}.

\bibitem[{He et~al.(2022)He, Chen, Xie, Li, Doll\'ar, and Girshick}]{he2021masked}
Kaiming He, Xinlei Chen, Saining Xie, Yanghao Li, Piotr Doll\'ar, and Ross Girshick. 2022.
\newblock Masked autoencoders are scalable vision learners.
\newblock In \emph{CVPR}, pages 16000--16009.

\bibitem[{Hochreiter and Schmidhuber(1997)}]{10.1162/neco.1997.9.8.1735}
Sepp Hochreiter and J{\"u}rgen Schmidhuber. 1997.
\newblock Long short-term memory.
\newblock \emph{Neural computation}, 9(8):1735--1780.

\bibitem[{Houlsby et~al.(2019)Houlsby, Giurgiu, Jastrzebski, Morrone, De~Laroussilhe, Gesmundo, Attariyan, and Gelly}]{houlsbyParameterEfficientTransferLearning2019}
Neil Houlsby, Andrei Giurgiu, Stanislaw Jastrzebski, Bruna Morrone, Quentin De~Laroussilhe, Andrea Gesmundo, Mona Attariyan, and Sylvain Gelly. 2019.
\newblock \href {https://proceedings.mlr.press/v97/houlsby19a.html} {Parameter-efficient transfer learning for {NLP}}.
\newblock In \emph{ICML}, pages 2790--2799.

\bibitem[{Jia et~al.(2021)Jia, Yang, Xia, Chen, Parekh, Pham, Le, Sung, Li, and Duerig}]{jiaScalingVisualVisionLanguage2021}
Chao Jia, Yinfei Yang, Ye~Xia, Yi-Ting Chen, Zarana Parekh, Hieu Pham, Quoc Le, Yun-Hsuan Sung, Zhen Li, and Tom Duerig. 2021.
\newblock \href {https://proceedings.mlr.press/v139/jia21b.html} {Scaling up visual and vision-language representation learning with noisy text supervision}.
\newblock In \emph{ICML}, pages 4904--4916.

\bibitem[{Kingma and Ba(2014)}]{kingma2017adam}
Diederik~P Kingma and Jimmy Ba. 2014.
\newblock Adam: A method for stochastic optimization.
\newblock \emph{arXiv preprint arXiv:1412.6980}.

\bibitem[{Krishna et~al.(2017)Krishna, Zhu, Groth, Johnson, Hata, Kravitz, Chen, Kalantidis, Li, Shamma et~al.}]{krishna2016visual}
Ranjay Krishna, Yuke Zhu, Oliver Groth, Justin Johnson, Kenji Hata, Joshua Kravitz, Stephanie Chen, Yannis Kalantidis, Li-Jia Li, David~A Shamma, et~al. 2017.
\newblock Visual genome: Connecting language and vision using crowdsourced dense image annotations.
\newblock \emph{IJCV}, 123:32--73.

\bibitem[{Krojer et~al.(2022)Krojer, Adlakha, Vineet, Goyal, Ponti, and Reddy}]{Krojer_2022}
Benno Krojer, Vaibhav Adlakha, Vibhav Vineet, Yash Goyal, Edoardo Ponti, and Siva Reddy. 2022.
\newblock Image retrieval from contextual descriptions.
\newblock \emph{arXiv preprint arXiv:2203.15867}.

\bibitem[{Kwon et~al.(2022)Kwon, Cai, Ravichandran, Bas, Bhotika, and Soatto}]{kwonMaskedVisionLanguage2023}
Gukyeong Kwon, Zhaowei Cai, Avinash Ravichandran, Erhan Bas, Rahul Bhotika, and Stefano Soatto. 2022.
\newblock Masked vision and language modeling for multi-modal representation learning.
\newblock \emph{arXiv preprint arXiv:2208.02131}.

\bibitem[{Lecun et~al.(1998)Lecun, Bottou, Bengio, and Haffner}]{lecun1998gradient}
Y.~Lecun, L.~Bottou, Y.~Bengio, and P.~Haffner. 1998.
\newblock \href {https://doi.org/10.1109/5.726791} {Gradient-based learning applied to document recognition}.
\newblock \emph{IEEE}, 86(11):2278--2324.

\bibitem[{Li et~al.(2023{\natexlab{a}})Li, Li, Savarese, and Hoi}]{li2023blip2}
Junnan Li, Dongxu Li, Silvio Savarese, and Steven Hoi. 2023{\natexlab{a}}.
\newblock Blip-2: Bootstrapping language-image pre-training with frozen image encoders and large language models.
\newblock \emph{arXiv preprint arXiv:2301.12597}.

\bibitem[{Li et~al.(2022)Li, Li, Xiong, and Hoi}]{liBLIPBootstrappingLanguageImage2022}
Junnan Li, Dongxu Li, Caiming Xiong, and Steven Hoi. 2022.
\newblock \href {https://proceedings.mlr.press/v162/li22n.html} {{BLIP}: Bootstrapping language-image pre-training for unified vision-language understanding and generation}.
\newblock In \emph{ICML}, pages 12888--12900.

\bibitem[{Li et~al.(2021)Li, Selvaraju, Gotmare, Joty, Xiong, and Hoi}]{liAlignFuseVision2021}
Junnan Li, Ramprasaath Selvaraju, Akhilesh Gotmare, Shafiq Joty, Caiming Xiong, and Steven Chu~Hong Hoi. 2021.
\newblock \href {https://proceedings.neurips.cc/paper_files/paper/2021/file/505259756244493872b7709a8a01b536-Paper.pdf} {Align before fuse: Vision and language representation learning with momentum distillation}.
\newblock In \emph{NeurIPS}, volume~34, pages 9694--9705.

\bibitem[{Li et~al.(2023{\natexlab{b}})Li, Hu, Ding, Ma, and Zhang}]{li2023neural}
Yunxin Li, Baotian Hu, Yunxin Ding, Lin Ma, and Min Zhang. 2023{\natexlab{b}}.
\newblock A neural divide-and-conquer reasoning framework for image retrieval from linguistically complex text.
\newblock \emph{arXiv preprint arXiv:2305.02265}.

\bibitem[{Lin et~al.(2014)Lin, Maire, Belongie, Hays, Perona, Ramanan, Doll{\'a}r, and Zitnick}]{lin2015microsoft}
Tsung-Yi Lin, Michael Maire, Serge Belongie, James Hays, Pietro Perona, Deva Ramanan, Piotr Doll{\'a}r, and C.~Lawrence Zitnick. 2014.
\newblock Microsoft coco: Common objects in context.
\newblock In \emph{ECCV}, pages 740--755.

\bibitem[{Lu et~al.(2023)Lu, Ding, Huo, Yang, Lu, Tomizuka, and Zhan}]{lu2023uniadapter}
Haoyu Lu, Mingyu Ding, Yuqi Huo, Guoxing Yang, Zhiwu Lu, Masayoshi Tomizuka, and Wei Zhan. 2023.
\newblock Uniadapter: Unified parameter-efficient transfer learning for cross-modal modeling.
\newblock \emph{arXiv preprint arXiv:2302.06605}.

\bibitem[{Lu et~al.(2019)Lu, Batra, Parikh, and Lee}]{luViLBERTPretrainingTaskAgnostic2019}
Jiasen Lu, Dhruv Batra, Devi Parikh, and Stefan Lee. 2019.
\newblock \href {https://proceedings.neurips.cc/paper_files/paper/2019/file/c74d97b01eae257e44aa9d5bade97baf-Paper.pdf} {Vilbert: Pretraining task-agnostic visiolinguistic representations for vision-and-language tasks}.
\newblock In \emph{NeurIPS}, volume~32.

\bibitem[{Plummer et~al.(2015)Plummer, Wang, Cervantes, Caicedo, Hockenmaier, and Lazebnik}]{plummer2016flickr30k}
Bryan~A. Plummer, Liwei Wang, Chris~M. Cervantes, Juan~C. Caicedo, Julia Hockenmaier, and Svetlana Lazebnik. 2015.
\newblock Flickr30k entities: Collecting region-to-phrase correspondences for richer image-to-sentence models.
\newblock In \emph{ICCV}.

\bibitem[{Radford et~al.(2021)Radford, Kim, Hallacy, Ramesh, Goh, Agarwal, Sastry, Askell, Mishkin, Clark, Krueger, and Sutskever}]{radford2021learning}
Alec Radford, Jong~Wook Kim, Chris Hallacy, Aditya Ramesh, Gabriel Goh, Sandhini Agarwal, Girish Sastry, Amanda Askell, Pamela Mishkin, Jack Clark, Gretchen Krueger, and Ilya Sutskever. 2021.
\newblock \href {https://proceedings.mlr.press/v139/radford21a.html} {Learning transferable visual models from natural language supervision}.
\newblock In \emph{ICML}, volume 139, pages 8748--8763.

\bibitem[{Sun et~al.(2023)Sun, Fang, Wu, Wang, and Cao}]{sun2023evaclip}
Quan Sun, Yuxin Fang, Ledell Wu, Xinlong Wang, and Yue Cao. 2023.
\newblock Eva-clip: Improved training techniques for clip at scale.
\newblock \emph{arXiv preprint arXiv:2303.15389}.

\bibitem[{Tong et~al.(2024)Tong, Liu, Zhai, Ma, LeCun, and Xie}]{tong2024eyes}
Shengbang Tong, Zhuang Liu, Yuexiang Zhai, Yi~Ma, Yann LeCun, and Saining Xie. 2024.
\newblock Eyes wide shut? exploring the visual shortcomings of multimodal llms.
\newblock \emph{arXiv preprint arXiv:2401.06209}.

\bibitem[{Wang et~al.(2023)Wang, Tang, Wang, Guo, Deng, and Han}]{wangMaskedImageModeling2023}
Haoqing Wang, Yehui Tang, Yunhe Wang, Jianyuan Guo, Zhi-Hong Deng, and Kai Han. 2023.
\newblock Masked image modeling with local multi-scale reconstruction.
\newblock In \emph{CVPR}, pages 2122--2131.

\bibitem[{Wang et~al.(2022)Wang, Yang, Men, Lin, Bai, Li, Ma, Zhou, Zhou, and Yang}]{wangOFAUnifyingArchitectures2022}
Peng Wang, An~Yang, Rui Men, Junyang Lin, Shuai Bai, Zhikang Li, Jianxin Ma, Chang Zhou, Jingren Zhou, and Hongxia Yang. 2022.
\newblock \href {https://proceedings.mlr.press/v162/wang22al.html} {{OFA}: Unifying architectures, tasks, and modalities through a simple sequence-to-sequence learning framework}.
\newblock In \emph{ICML}, volume 162, pages 23318--23340.

\bibitem[{Xu et~al.(2023)Xu, Xie, Tan, Huang, Howes, Sharma, Li, Ghosh, Zettlemoyer, and Feichtenhofer}]{xu2023demystifying}
Hu~Xu, Saining Xie, Xiaoqing~Ellen Tan, Po-Yao Huang, Russell Howes, Vasu Sharma, Shang-Wen Li, Gargi Ghosh, Luke Zettlemoyer, and Christoph Feichtenhofer. 2023.
\newblock Demystifying clip data.
\newblock \emph{arXiv preprint arXiv:2309.16671}.

\bibitem[{Yang et~al.(2023)Yang, Li, Lin, Wang, Lin, Liu, and Wang}]{yang2023dawn}
Zhengyuan Yang, Linjie Li, Kevin Lin, Jianfeng Wang, Chung-Ching Lin, Zicheng Liu, and Lijuan Wang. 2023.
\newblock The dawn of lmms: Preliminary explorations with gpt-4v (ision).
\newblock \emph{arXiv preprint arXiv:2309.17421}, 9(1):1.

\bibitem[{Zhai et~al.(2023)Zhai, Mustafa, Kolesnikov, and Beyer}]{zhai2023sigmoid}
Xiaohua Zhai, Basil Mustafa, Alexander Kolesnikov, and Lucas Beyer. 2023.
\newblock Sigmoid loss for language image pre-training.
\newblock \emph{arXiv preprint arXiv:2303.15343}.

\bibitem[{Zhang et~al.(2021)Zhang, Fang, Zhang, Gao, Li, Dai, Qiao, and Li}]{zhangTipAdapterTrainingfreeCLIPAdapter2021}
Renrui Zhang, Rongyao Fang, Wei Zhang, Peng Gao, Kunchang Li, Jifeng Dai, Yu~Qiao, and Hongsheng Li. 2021.
\newblock Tip-{{Adapter}}: {{Training-free CLIP-Adapter}} for {{Better Vision-Language Modeling}}.

\end{thebibliography}

\appendix

%\section{Appendix}
\appendix

\section{Implementation details}
\label{apx:implementation}

\subsection{Training Hyper-parameters}
\label{apx:1}
\textbf{Pre-training.}
We conducted pre-training on 4 Nvidia A100 GPUs for about 10 hours, randomly sampling $224^2$ patches from images and using RandAugment for data augmentation.

\begin{table}[h]
\centering
\small
\centering\setlength{\tabcolsep}{2pt}
\begin{tabular}{l|l}
\toprule
Hyperparameters        &              \\ \midrule
max epoch             & 20           \\
batch size             & 256          \\
vit                   & Vit-B/16     \\
input resolution       & $224^2$    \\
augmentation           & RandAug      \\
optimizer              & AdamW        \\
base learning rate     & 3e-4         \\
warmup learning rate   & 1e-6         \\
minimize learning rate & 1e-6         \\
momentum               & (0.9, 0.999) \\
warmup steps           & 3000         \\
weight decay           & 1e-4         \\
mask ratio             & 0.25         \\
reduction              & 2            \\
random seed            & 42           \\ \bottomrule
\end{tabular}
\caption{Pretraining setting on IMAGECODE}
\label{tab:pretrain}
\end{table}
\noindent
\textbf{Fine-tuning.}
Compared to Krojer \etal's multi-step approach involving +Context Batch, +Context Module, and +Temporal Embeddings, we streamline Krojer \etal's procedure by directly fine-tuning our model on IMAGECODE and then separately training a contextual modeling module between the backbone and the prediction head to enable contextual comparison and reasoning. 

During the first stage, we conduct full fine-tuning for 20 epochs. After that, we add a two-layer transformer as a context/temporal module, which is identical to Krojer \etal. To optimize the module while keeping other parameters fixed, we freeze all parts of the model except the inserted module and train it alone for 5 epochs. 

We set the batch size to 360 images (36 batches of 10-image-set) and use the Adam~\cite{kingma2017adam} optimizer with learning rate of 2e-6 and weight decay of 0.01. We conduct the experiment on one NVIDIA Geforce 3090 for about one day.  All other baselines also adopted the same settings.

\begin{table}[h]
\centering
\small
\centering\setlength{\tabcolsep}{2pt}
\begin{tabular}{l|l}
\toprule
Hyperparameters         &                     \\ \midrule
max epoch               & 25                  \\
batch size              & 36                  \\
input resolution        & $224^2$           \\
augmentation            & RandAug             \\
optimizer               & Adam                \\
backbone learning rate  & 2e-6                \\
head learning rate      & 1e-4                \\
momentum                & (0.9, 0.999)        \\
learning rate scheduler & ExponentialLR(0.95) \\
weight decay            & 1e-2                \\
random seed             & 10                   \\ \bottomrule
\end{tabular}
\caption{Fine-tuning setting on IMAGECODE}
\label{tab:finetune}
\end{table}

\subsection{Dataset Details}

Table \ref{tab:dataset} provides statistics for the datasets used in pre-training and fine-tuning. It includes three datasets: COCO, VG (Visual Genome), and IMAGECODE. The number of images (\#image) and texts (\#text) for each dataset are listed.
\begin{table}[h]
\centering
\small
\centering\setlength{\tabcolsep}{2pt}
\begin{tabular}{l|ccc}
\toprule
        & COCO & VG & IMAGECODE \\ \midrule
\#image & 113K          & 100K        & 94020              \\
\#text  & 567K          & 769K        & 21202              \\ \bottomrule
\end{tabular}
\caption{Details of datasets.}
\label{tab:dataset}
\end{table}

\subsection{Model Details}
\label{apx:model}

In section 3.2, the scorer is a 1x768 linear layer. In section 3.3 (Step 2), the decoder is a four-layer transformer, with each layer having twelve attention heads and a feature dimension of 768.

In section 4.2, for CLIP, ViLBERT, and UNITER, we utilized the models with added context modules and temporal embeddings as proposed by Krojer et al. Our-Context model incorporated a context module in the same manner, while other models did not have such a module added. For ALBEF, due to its pretraining on images with a resolution of 256x256, we conducted experiments using a slightly higher resolution (224x224) compared to other models.

\section{Validation Performance}
\label{apx:B}
Comparison with state-of-the-art methods on the valid set of IMAGECODE is shown on Table \ref{tab:validation11}.

\begin{table*}[]
\centering

\begin{tabular}{lccccccc}
\toprule
\multicolumn{1}{c}{}       & \textbf{}       & \multicolumn{3}{c}{Zero-shot}          & \multicolumn{3}{c}{Fine-tune}          \\ \cmidrule(r){3-5}  \cmidrule(r){6-8}
                           Method& Params & All  & Video & Image & All  & Video & Image \\ \midrule
CLIP~\cite{radford2021learning}              & 473M            & 22.4          & 15.6           & 47.8           & 29.9          & 22.0           & 59.8           \\
UNITER~\cite{chenUNITERUNiversalImageTExt2020}            & -               & 19.8          & 13.6           & 42.9           & 25.7          & 19.1           & 50.5           \\
ViLBERT~\cite{luViLBERTPretrainingTaskAgnostic2019}           & -               & 19.3          & 13.5           & 40.8           & 24.5          & 18.0           & 49.3           \\
OFA~\cite{wangOFAUnifyingArchitectures2022}               & -               & -             & -              & -              & 27.2          & 21.0           & 52.1           \\
ALBEF~\cite{liAlignFuseVision2021}             & -               & 27.7          & 15.7           & 73.3           & -             & -              & -              \\
M3AE~\cite{gengMultimodalMaskedAutoencoders2022}              & -               & -             & -              & -              & -             & -              & -              \\ 
BLIP~\cite{liBLIPBootstrappingLanguageImage2022}              & 223M            & 28.1          & 15.9           & 74.4           & \underline{34.1}          & 22.7           & \underline{77.4}           \\
BLIP-2~\cite{li2023blip2}            & 1.2B            & \underline{29.4}          & \underline{16.3}           & \textbf{79.2}  & -             & \textbf{-}     & \textbf{-}     \\
NDCR~\cite{li2023neural}              & 440M            & \textbf{-}    & \textbf{-}     & -              & \underline{34.1}          & \textbf{26.1}  & 64.3           \\ \midrule
ContextBLIP (Ours)              & 240M            & \textbf{31.0} & \textbf{18.8}  & \underline{77.1}           & \textbf{35.7}          & \underline{24.4}           & \textbf{78.5}  \\
\midrule
Human Performance & \multicolumn{7}{c}{\textbf{90.8}}                                                                                   \\ \bottomrule
\end{tabular}
\caption{Comparison with state-of-the-art methods on IMAGECODE task. Our model achieve highest zero-shot and fine-tune performance while requiring a relatively fewer number of parameters. The best and second-best results are marked in \textbf{bold} and \underline{underline}.}
\label{tab:main-results11}
\end{table*}

\begin{table*}[h]
\centering
\begin{tabular}{lccccccc}
\toprule
                      & \textbf{}       & \multicolumn{3}{c}{\textbf{Zero-shot}}          & \multicolumn{3}{c}{\textbf{Fine-tune}}          \\ \cmidrule(r){3-5}  \cmidrule(r){6-8}
                     Method & Params & All  & Video & Image & All  & Video & Image \\ \midrule
CLIP~\cite{radford2021learning}         & 473M            & 21.8          & 14.9           & 51.6           & 30.6          & 22.3           & 67.0           \\
UNITER~\cite{chenUNITERUNiversalImageTExt2020}        & -               & 19.8          & 13.6           & 42.9           & 26.0          & 19.9           & 52.8           \\
ViLBERT~\cite{luViLBERTPretrainingTaskAgnostic2019}      & -               & 18.5          & 14.0           & 37.9           & 25.1          & 19.4           & 49.5           \\
ALBEF~\cite{liAlignFuseVision2021}        & -               & 28.2          & \underline{17.0}           & \underline{77.0}           & -             & -              & -              \\
OFA~\cite{wangOFAUnifyingArchitectures2022}          & -               & -             & -              & -              & 27.2          & 21.0           & 52.1           \\
BLIP~\cite{liBLIPBootstrappingLanguageImage2022}         & 223M            & \underline{28.4}          & \underline{17.0}           & \textbf{77.9}  & \underline{36.2}           & \underline{26.3}           & \underline{79.3}           \\
ContextBLIP (Ours)         & 240M            & \textbf{30.6} & \textbf{19.9}  & \underline{77.0}           & \textbf{38.5}          & \textbf{28.7}           & \textbf{81.2}  \\ \bottomrule
\end{tabular}
\caption{Validation Performance on IMAGECODE}
\label{tab:validation11}
\end{table*}

\begin{table*}[]
\centering
\small
\centering\setlength{\tabcolsep}{2pt}
\footnotesize
\begin{tabular}{l|cc|ccccccccc|c}
\toprule
                   & \makecell[c]{Image\\Size} & Params  & \includegraphics[width=0.03\textwidth]{imgs/table/1.pdf}    & \includegraphics[width=0.03\textwidth]{imgs/table/2.pdf}    & \includegraphics[width=0.03\textwidth]{imgs/table/3.pdf}    & \includegraphics[width=0.03\textwidth]{imgs/table/4.pdf}   & \includegraphics[width=0.018\textwidth]{imgs/table/5.pdf}    & \includegraphics[width=0.03\textwidth]{imgs/table/6.pdf}    & \includegraphics[width=0.04\textwidth]{imgs/table/7.pdf}    & \includegraphics[width=0.03\textwidth]{imgs/table/8.pdf}    & \includegraphics[width=0.03\textwidth]{imgs/table/9.pdf}    & \makecell[c]{MMVP\\Average} \\ \midrule
\makecell[c]{OpenAI ViT-L-14\\\cite{radford2021learning}}    & $224^2$                                       & 427.6M  & 13.3 & 13.3 & 20.0 & 20.0 & 13.3 & 53.3 & 20.0 & 6.7  & 13.3 & 19.3                                        \\
\makecell[c]{OpenAI ViT-L-14\\\cite{radford2021learning}}    & $336^2$                                       & 427.9M  & 0.0  & 20.0 & 40.0 & 20.0 & 6.7  & 20.0 & 33.3 & 6.7  & 33.3 & 20.0                                        \\
\makecell[c]{SigLIP ViT-SO-14\\\cite{zhai2023sigmoid}}   & $224^2$                                       & 877.4M  & \textbf{26.7} & 20.0 & 53.3 & \textbf{40.0} & 20.0 & \underline{66.7} & \underline{40.0} & \underline{20.0} & \textbf{53.3} & 37.8                                        \\
\makecell[c]{SigLIP ViT-SO-14\\\cite{zhai2023sigmoid}}    & $384^2$                                       & 878.0M  & \underline{20.0} & \underline{26.7} & \underline{60.0} & \underline{33.3} & 13.3 & \underline{66.7} & 33.3 & \textbf{26.7} & \textbf{53.3} & 37.0                                        \\
\makecell[c]{DFN ViT-H-14\\\cite{fang2023data}}       & $224^2$                                       & 986.1M  & \underline{20.0} & \underline{26.7} & 73.3 & 26.7 & 26.7 & \underline{66.7} & \textbf{46.7} & 13.3 & \textbf{53.3} & \underline{39.3}                                        \\
DFN ViT-H-14       & $378^2$                                       & 986.7M  & 13.3 & 20.0 & 53.3 & \underline{33.3} & 26.7 & \underline{66.7} & \underline{40.0} & \underline{20.0} & 40.0 & 34.8                                        \\
\makecell[c]{MetaCLIP ViT-L-14\\\cite{xu2023demystifying}}  & $224^2$                                       & 427.6M  & 13.3 & 6.7  & \textbf{66.7} & 6.7  & \underline{33.3}  & 46.7 & 20.0 & 6.7  & 13.3 & 23.7                                        \\
\makecell[c]{MetaCLIP ViT-H-14\\\cite{xu2023demystifying}}  & $224^2$                                       & 986.1M  & 6.7  & 13.3 & \underline{60.0} & 13.3 & 6.7  & 53.3 & 26.7 & 13.3 & 33.3 & 25.2                                        \\
\makecell[c]{EVA01 ViT-g-14\\\cite{sun2023evaclip}}     & $224^2$                                       & 1136.4M & 6.7  & \underline{26.7} & 40.0 & 6.7  & 13.3 & \underline{66.7} & 13.3 & 13.3 & 20.0 & 23.0                                        \\
\makecell[c]{EVA02 ViT-bigE-14+\\\cite{sun2023evaclip}} & $224^2$                                       & 5044.9M & 13.3 & 20.0 & \textbf{66.7} & 13.3 & 26.7 & \underline{66.7} & 26.7 & \underline{20.0} & 33.3 & 33.3                                        \\
BLIP~\cite{liBLIPBootstrappingLanguageImage2022} &$224^2$  & 223M & 13.3 & 6.7 & 40.0 & 20.0 & 26.7 & \underline{66.7} & \textbf{46.7} & \underline{20.0} & \underline{46.7} & 31.9 \\ \midrule
ContextBLIP (Ours)               & $224^2$                                       & \textbf{240M}    & \textbf{26.7} & \textbf{46.7}  & \underline{60.0} & \textbf{40.0} & \textbf{46.7} & \textbf{93.3} & \underline{40.0} & \underline{20.0} & \textbf{53.3} & \textbf{47.4}    \\ \bottomrule

\end{tabular}
\caption{Comparison with various VLMs on different visual patterns in MMVP-VLM benchmark. The best and second-best results are marked in \textbf{bold} and \underline{underline}. We identify 9 visual patterns: \includegraphics[width=0.05\columnwidth]{imgs//table/1.pdf} : Orientation and Direction, \includegraphics[width=0.05\columnwidth]{imgs//table/2.pdf} : Presence of Specific Features, \includegraphics[width=0.05\columnwidth]{imgs//table/3.pdf} : State and Condition, \includegraphics[width=0.05\columnwidth]{imgs//table/4.pdf} : Quantity and Count,\includegraphics[width=0.03\columnwidth]{imgs//table/5.pdf} : Positional and Relational Context, \includegraphics[width=0.05\columnwidth]{imgs//table/6.pdf} : Color and Appearance, \includegraphics[width=0.05\columnwidth]{imgs//table/7.pdf} : Structural and Physical Characteristics, \includegraphics[width=0.04\columnwidth]{imgs//table/8.pdf} : Text and \includegraphics[width=0.05\columnwidth]{imgs//table/9.pdf} : Viewpoint and Perspective. }
\label{tab:MMVP11}
\end{table*}

\section{Additional Case Studies}

For tasks involving high image similarity and detailed textual descriptions, such as contrastive image retrieval, the challenge is considerable.    We will demonstrate the superior performance of our MIM adaption method in more scenarios and provide the following examples to further illustrate the doubly contextual alignment capabilities of our model.

\begin{figure*}[!h]
    \centering
    \includegraphics[height=0.48\textheight, width=2\columnwidth]{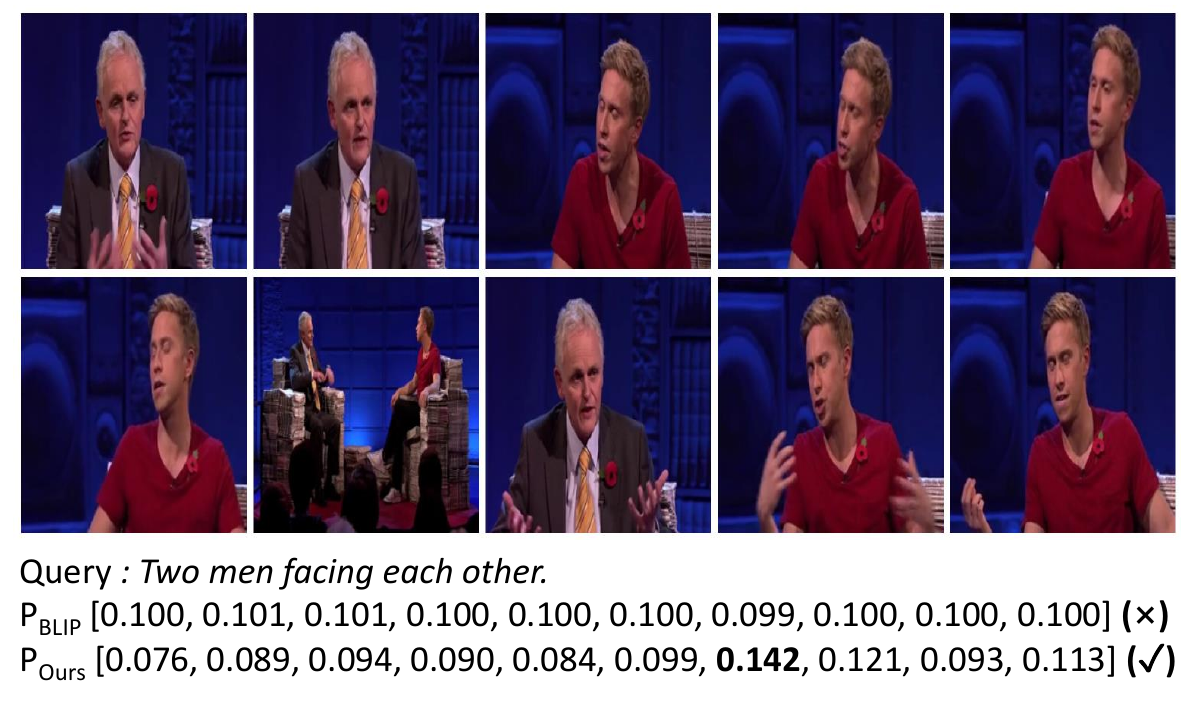}
    \includegraphics[height=0.48\textheight, width=2\columnwidth]{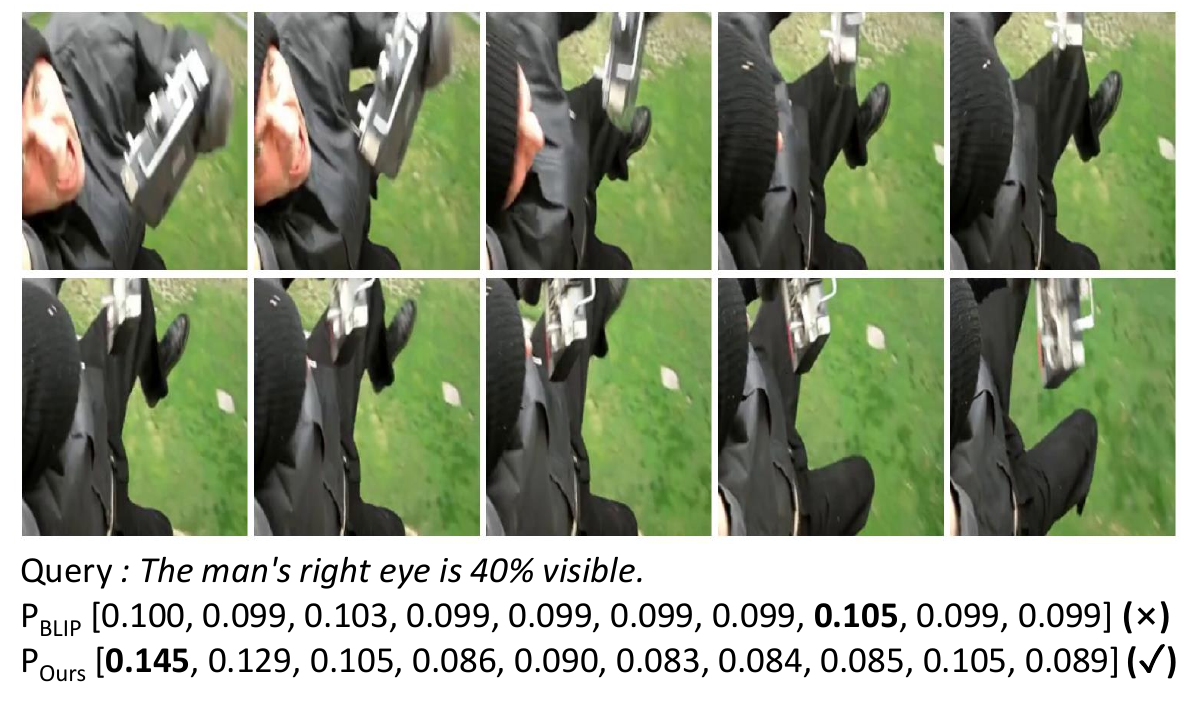}
    \caption{Zero-shot cases from the test set. Our model has advantages over BLIP in both confidence scores.}
    \label{fig:zeroshot_case_apx}
\end{figure*}

Figure \ref{fig:zeroshot_case_apx} presents examples of two zero-shot experiments.    In the left image, when two men are facing each other, our model assigns significantly higher matching scores compared to other images.    In the right image, despite the difficulty in estimating the proportion of the man's right eye visible, our model assigns obviously higher scores to the first two images showing the right eye.    In contrast, BLIP assigns similar scores to all ten images in both samples, indicating its difficulty in attending to the textual cues used to align the intra-contexual information.

\begin{figure*}[!h]
    \centering
    \includegraphics[height=0.48\textheight, width=2\columnwidth]{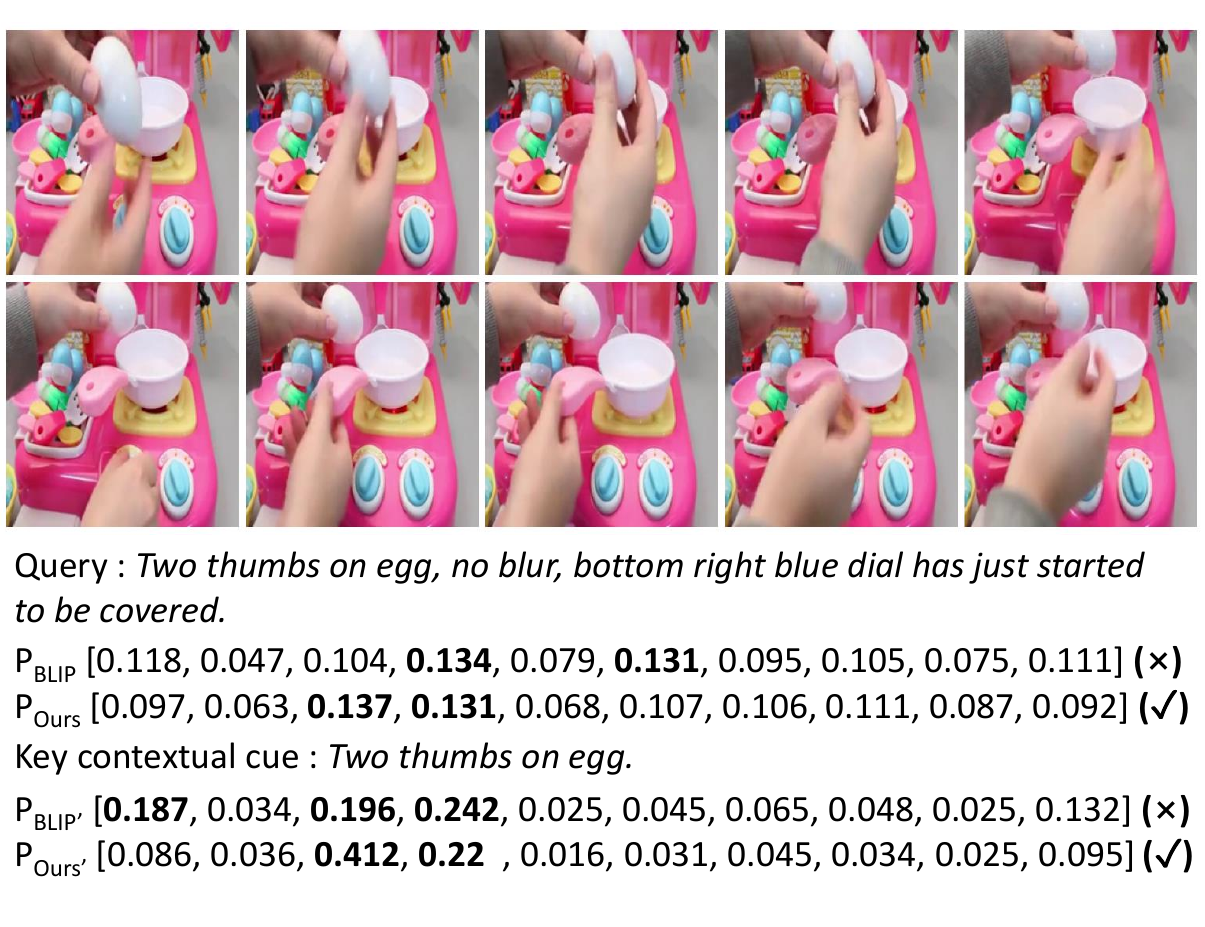}
    \includegraphics[height=0.48\textheight, width=2\columnwidth]{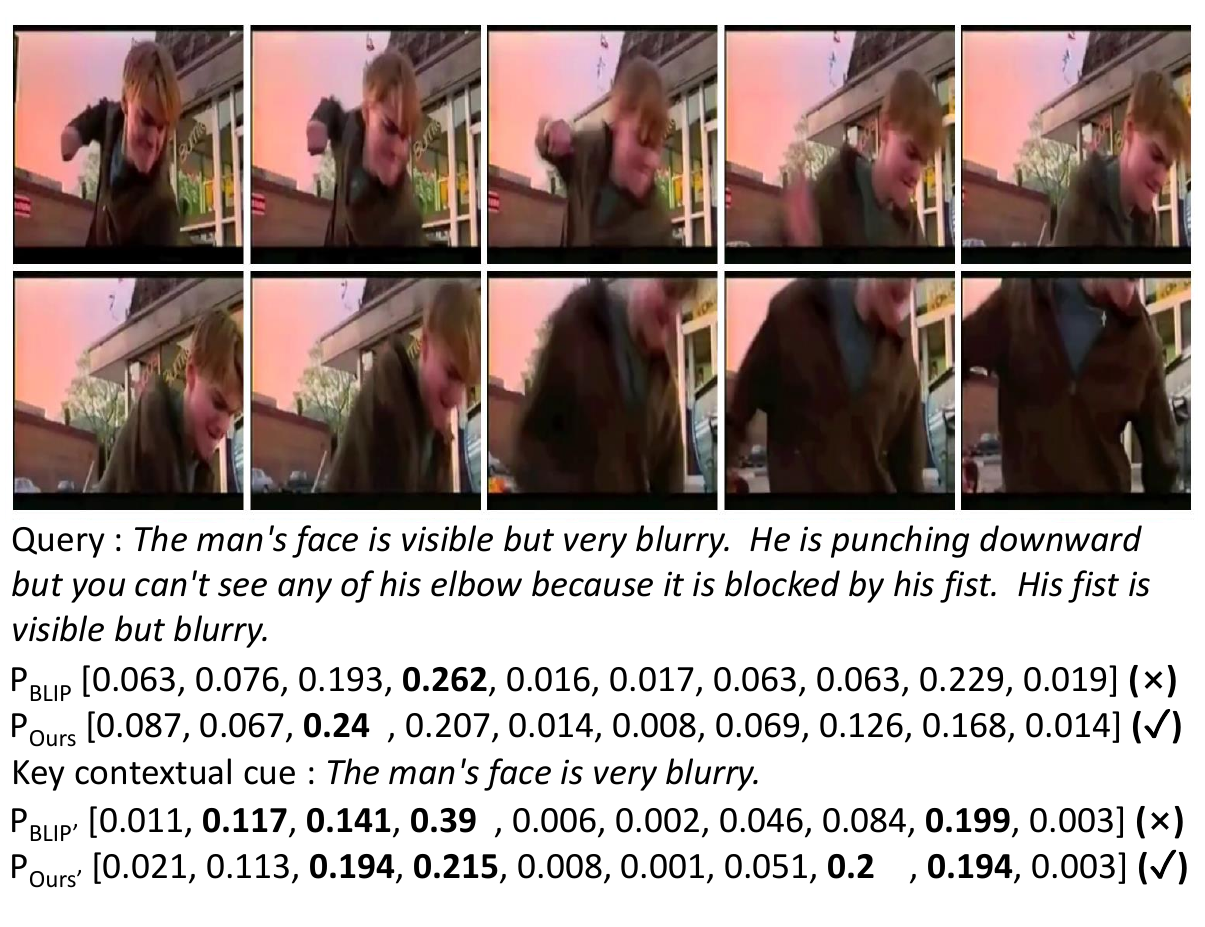}
    \caption{Two cases from the test set after fine-tuning. Our model outperforms BLIP in confidence scores for both compound and simple proposition texts.}
    \label{fig:finetune_case_apx}
\end{figure*}

Figure \ref{fig:finetune_case_apx} showcases examples of fine-tuning experiments.   The left image in Figure \ref{fig:finetune_case_apx} illustrates an example of the Quantity phenomenon, with the subheading "Two thumbs on the egg."    We observe that our model infers significantly higher confidence scores for images 2 and 3 (matching the description of thumb quantity) and infers higher BLIP confidence scores for images 0, 2, and 3 (where the description of thumb quantity does not match).    Thus, we demonstrate that our model indeed performs better in dual-contexual alignment.

The right image in Figure \ref{fig:finetune_case_apx} demonstrates an example of the Meta-property phenomenon, with the subheading "The man's face is blurry."    We find that our model infers higher confidence scores for images 2, 3, 7, and 8 (all relatively matching the description) while BLIP infers higher confidence scores for images 1, 2, 3, and 8 (where image 1 is clear but does not match the description).    Hence, it indicates that our model indeed performs better in understanding intra-contexual alignment.

In conjunction with the main text, we demonstrate the advantages of our model in two challenges: better intra-contexual alignment, better inter-contexual alignment. It can be seen that in many scenarios, our model exhibits better dual-contextual alignment capabilities compared to BLIP.

%Additional Case Studies的两个插图

\section{More about GPT-4V Experiment}

\subsection{Sampling Details}

\begin{figure*}[!h]
    \centering
    \includegraphics[scale=0.8]{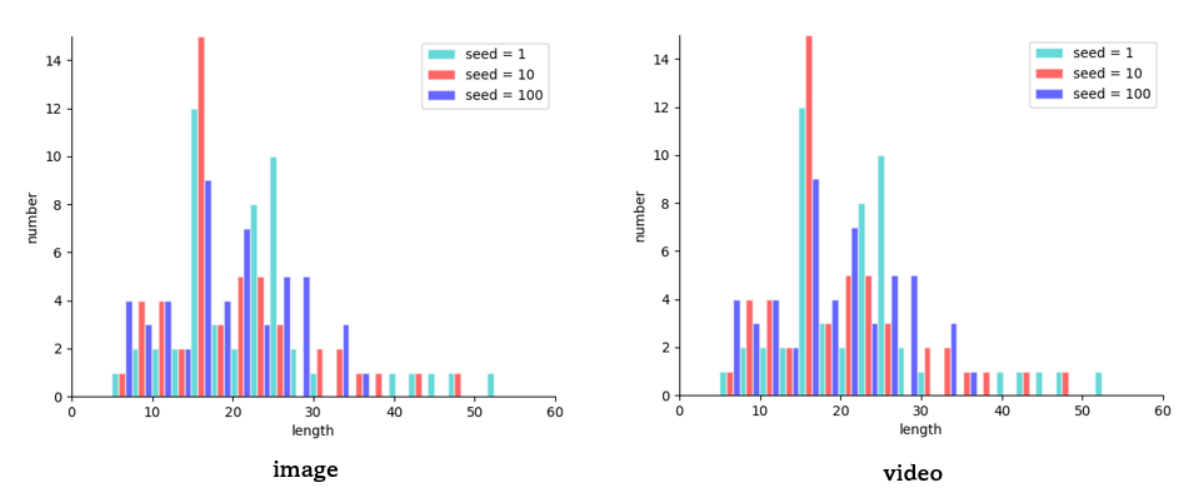}
    \caption{Distribution of the number of tokens across contextual descriptions in the subset of IMAGECODE. The distributions of the three sampling results are similar to the overall distribution described in the IMAGECODE text, and the average text length is about 20.}
    \label{fig:dist}
\end{figure*}

Due to GPT-4V's high computational demands, we created distinct test datasets from IMAGECODE's static and video-shot sections. We randomly chose 50 samples from each, repeating thrice to avoid bias. Each sample included a ground-truth image plus nine adjacent ones, since GPT-4V processes up to ten images simultaneously. We prompted GPT-4V in form of "Which image fits the text description best? Please output the serial number of the image:$<text\_query>$" to select the matching image.  We used three random seeds: 1, 10, 100, to explore the datasets with the text length distribution (in tokens) illustrated in the Figure \ref{fig:dist}.

In this experiment, using Python scripts with the API interface of the gpt-4-vision-preview model, we make a request to GPT-4V. If encounter a refusal to answer, we use the browser version to ask again until a result is obtained.

\subsection{More Comparison with GPT-4V}
\label{apx:D2}

This part provides some more examples of GPT4V prediction errors, where our model predicts correctly. The first two samples are from video shot set, and the last two are from static pictures.
%修改前两个和后两个的表述，改为图片的名称

\begin{figure*}[!ht] 
    \centering
    \includegraphics[width=\textwidth]{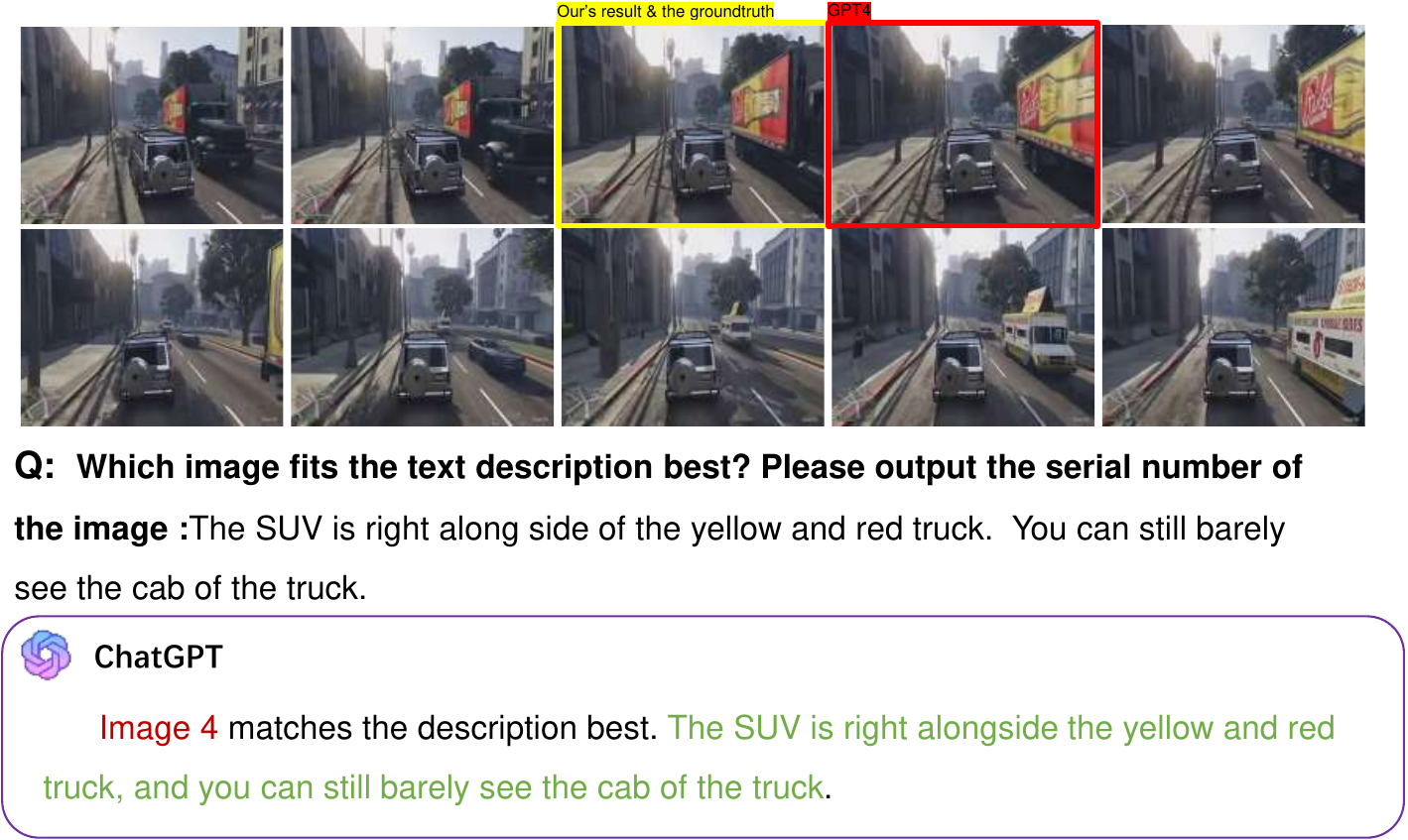} % 将宽度设置为整个文本的宽度
    \caption{A case from the video-shot set. This is an enlarged version of the example in the main text. 
    The yellow boxes indicate the correct image and our model's result, while the red boxes represent the model's output. We will see our model's result is correct but the GPT-4V's output is wrong.}
    \label{fig:mylabel1}%修改label
\end{figure*}

\begin{figure*}[!ht] 
    \centering
    \includegraphics[width=\textwidth]{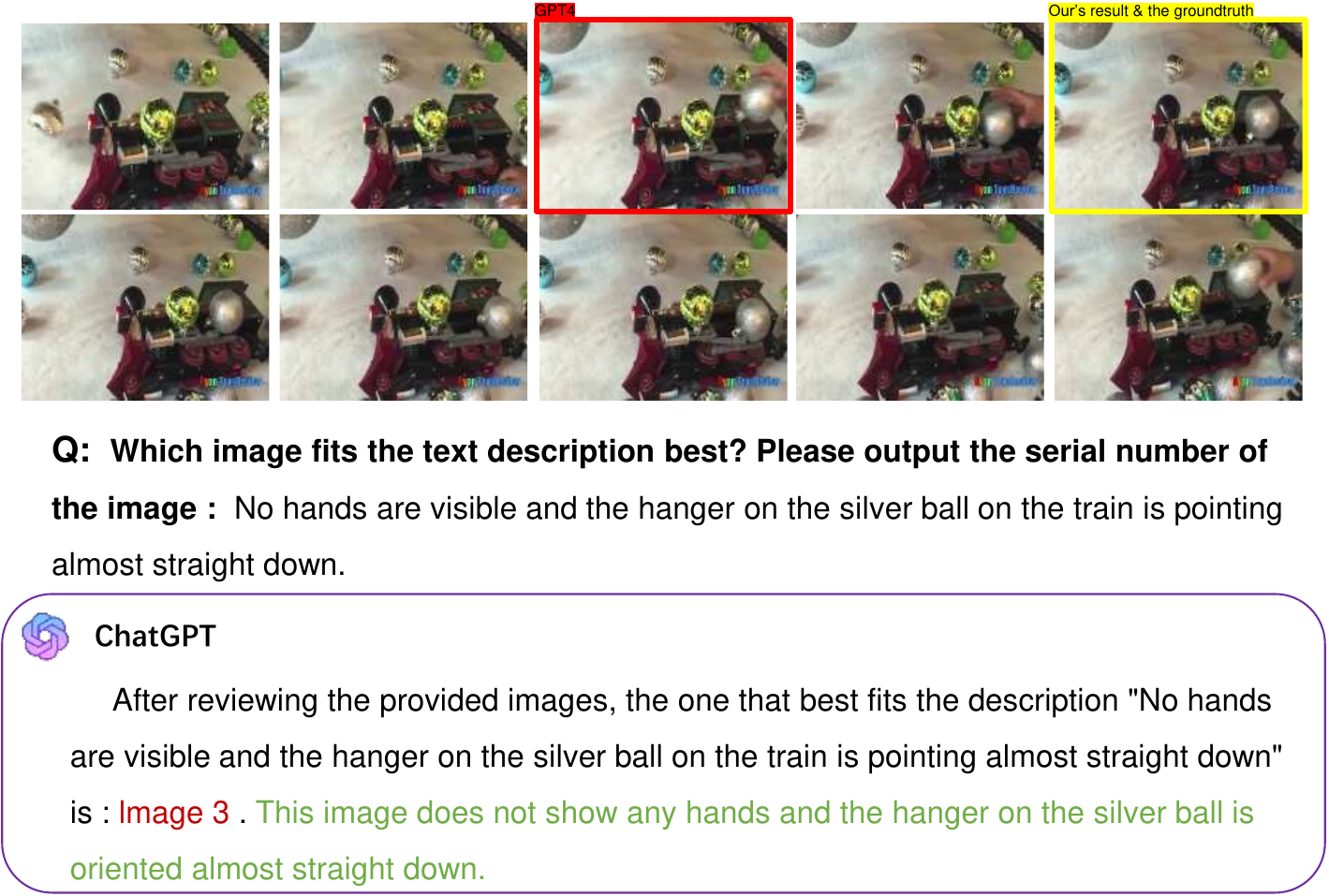} % 将宽度设置为整个文本的宽度
    \caption{A case from the video-shot set.
    The yellow boxes indicate the correct image and our model's result, while the red boxes represent the model's output. We will see our model's result is correct but the GPT-4V's output is wrong.}
    \label{fig:mylabel2}%修改label
\end{figure*}

\begin{figure*}[!ht] 
    \centering
    \includegraphics[width=\textwidth]{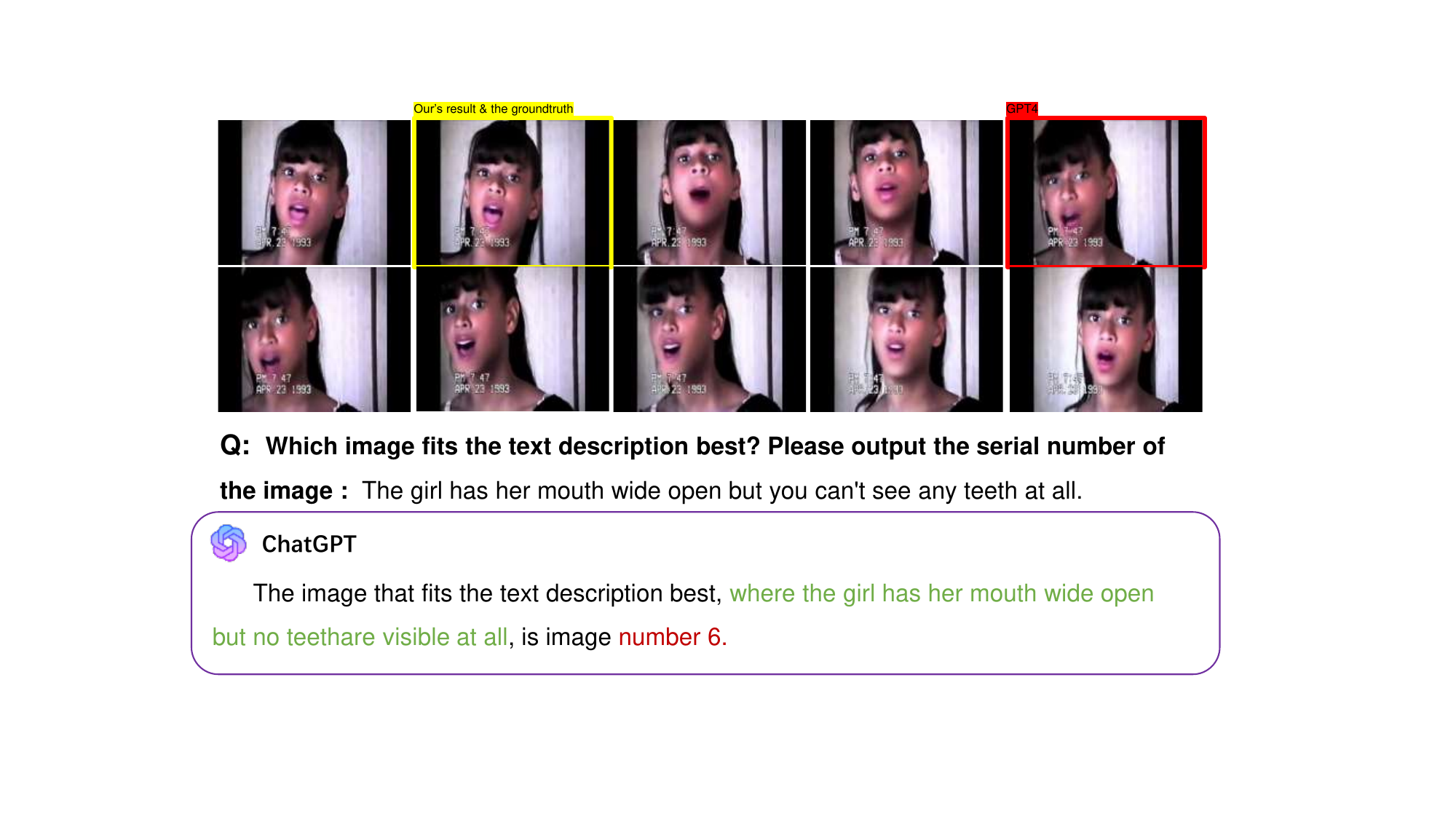} % 将宽度设置为整个文本的宽度
    \caption{A case from the video-shot set.
    The yellow boxes indicate the correct image and our model's result, while the red boxes represent the model's output. We will see our model's result is correct but the GPT-4V's output is wrong.}
    \label{fig:mylabel3}%修改label
\end{figure*}

\begin{figure*}[!ht] 
    \centering
    \includegraphics[width=\textwidth]{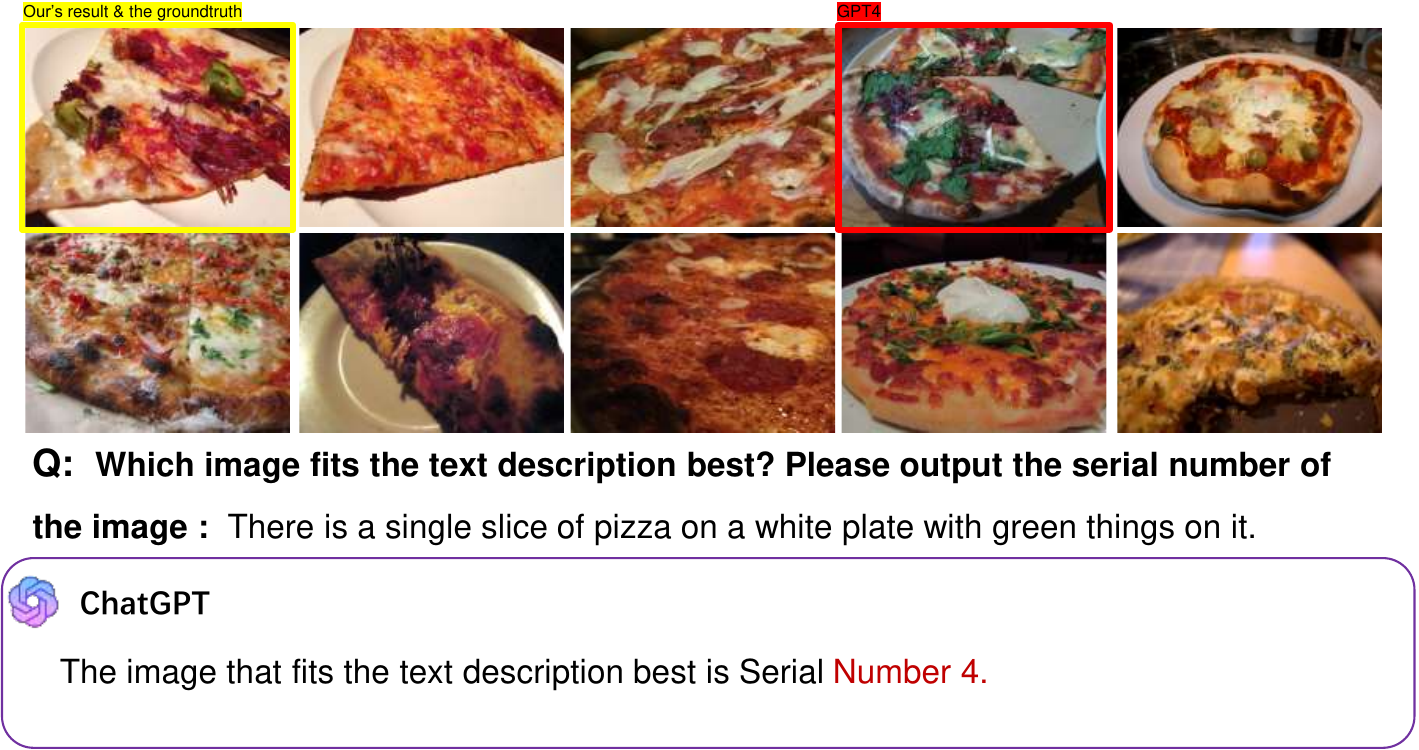} % 将宽度设置为整个文本的宽度
    \caption{A case from the static pictures set.
    The yellow boxes indicate the correct image and our model's result, while the red boxes represent the model's output. We will see our model's result is correct but the GPT-4V's output is wrong.}
    \label{fig:mylabel4}%修改label
\end{figure*}

\begin{figure*}[!ht] 
    \centering
    \includegraphics[width=\textwidth]{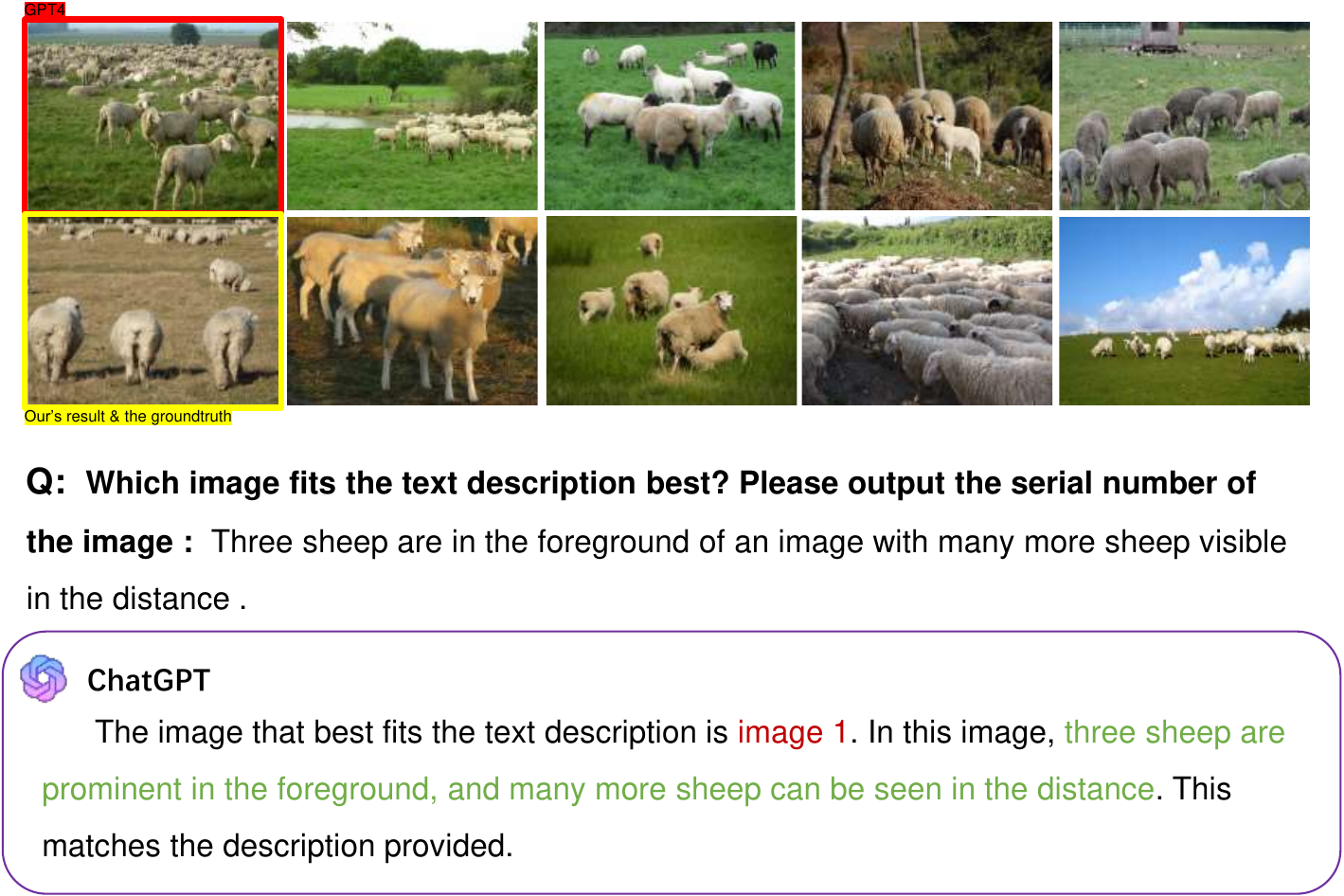} % 将宽度设置为整个文本的宽度
    \caption{A case from the static pictures set.
    The yellow boxes indicate the correct image and our model's result, while the red boxes represent the model's output. We will see our model's result is correct but the GPT-4V's output is wrong.}
    \label{fig:mylabel5}%修改label
\end{figure*}

\end{document}